\documentclass[preprint,12pt]{elsarticle}

\usepackage{amssymb}
\usepackage{amsmath}
\usepackage{lineno}
\usepackage{hyperref}
\usepackage{url}
\usepackage{booktabs}       % professional-quality tables
\usepackage{amsfonts}       % blackboard math symbols
\usepackage{nicefrac}       % compact symbols for 1/2, etc.
\usepackage{xcolor}         % colors
\usepackage{graphicx}
\usepackage{wrapfig}
\usepackage{multirow}
\usepackage{epsfig}
\usepackage{natbib}
\usepackage{amsmath}
\usepackage{amssymb}
\usepackage{subcaption}
\usepackage{marvosym}
\usepackage{color}
\usepackage{threeparttable}
\usepackage{algorithm}
\usepackage{algpseudocode}
\usepackage{setspace}
\usepackage{amsthm}
\usepackage{tikz}
\usetikzlibrary{arrows, automata, positioning}
\usepackage[xcolor]{changes}

\usepackage{pifont}% http://ctan.org/pkg/pifont
\newcommand{\update}[1]{{\textcolor{black}{#1}}}

\newtheorem{theorem}{Theorem}

\newcommand{\cmark}{\ding{51}}%
\newcommand{\xmark}{\ding{55}}%
\usepackage[xcolor]{changes}

\algrenewcommand\algorithmicrequire{\textbf{Input:}}
\algrenewcommand\algorithmicensure{\textbf{Output:}}
\journal{Information Fusion}

\begin{document}

\begin{frontmatter}

%% Title, authors and addresses

%% use the tnoteref command within \title for footnotes;
%% use the tnotetext command for theassociated footnote;
%% use the fnref command within \author or \affiliation for footnotes;
%% use the fntext command for theassociated footnote;
%% use the corref command within \author for corresponding author footnotes;
%% use the cortext command for theassociated footnote;
%% use the ead command for the email address,
%% and the form \ead[url] for the home page:
%% \title{Title\tnoteref{label1}}
%% \tnotetext[label1]{}
%% \author{Name\corref{cor1}\fnref{label2}}
%% \ead{email address}
%% \ead[url]{home page}
%% \fntext[label2]{}
%% \cortext[cor1]{}
%% \affiliation{organization={},
%%             addressline={},
%%             city={},
%%             postcode={},
%%             state={},
%%             country={}}
%% \fntext[label3]{}

\title{TSCMamba: Mamba Meets Multi-View Learning for Time Series Classification}

\author[cs]{Md Atik Ahamed}
\ead{atikahamed@uky.edu}

\author[cs,ibi]{Qiang Cheng\corref{cor1}}
\cortext[cor1]{Corresponding author}
\ead{qiang.cheng@uky.edu}

\affiliation[cs]{organization={Department of Computer Science, University of Kentucky},
            city={Lexington},
            state={KY},
            country={USA}}

\affiliation[ibi]{organization={Institute for Biomedical Informatics, University of Kentucky},
            city={Lexington},
            state={KY},
            country={USA}}

\date{}
%% Abstract
\begin{abstract}
Multivariate time series classification (TSC) is critical for various applications in fields such as healthcare and finance. While various approaches for TSC have been explored, important properties of time series, such as shift equivariance and inversion invariance, are largely underexplored by existing works. To fill this gap, we propose a novel multi-view approach to capture patterns with properties like shift equivariance. Our method integrates diverse features, including spectral, temporal, local, and global features, to obtain rich, complementary contexts for TSC. We use continuous wavelet transform to capture time-frequency features that remain consistent even when the input is shifted in time. These features are fused with temporal convolutional or multilayer perceptron features to provide complex local and global contextual information. We utilize the Mamba state space model for efficient and scalable sequence modeling and capturing long-range dependencies in time series. Moreover, we introduce a new scanning scheme for Mamba, called tango scanning, to effectively model sequence relationships and leverage inversion invariance, thereby enhancing our model's generalization and robustness. Experiments on two sets of benchmark datasets (10+20 datasets) demonstrate our approach's effectiveness, achieving average accuracy improvements of 4.01-6.45\% and 7.93\% respectively, over leading TSC models such as TimesNet and TSLANet.\\
\end{abstract}

% %%Graphical abstract
% \begin{graphicalabstract}
% %\includegraphics{grabs}
% \end{graphicalabstract}

% %%Research highlights
% \begin{highlights}
% \item Research highlight 1
% \item Research highlight 2
% \end{highlights}

%% Keywords
\begin{keyword}
%% keywords here, in the form: keyword \sep keyword
Time Series Classification\sep Deep Learning \sep state-space-machine \sep multi-view learning
%% PACS codes here, in the form: \PACS code \sep code

%% MSC codes here, in the form: \MSC code \sep code
%% or \MSC[2008] code \sep code (2000 is the default)

\end{keyword}

\end{frontmatter}

%% Add \usepackage{lineno} before \begin{document} and uncomment 
%% following line to enable line numbers
% \linenumbers

\section{Introduction}
Time series classification (TSC) is a fundamental task in diverse fields abundant with time series data such as healthcare, weather forecasting, and finance. With the advancement of sensing technologies, multivariate time series (MTS) data have been ubiquitous, and thus TSC over MTS has attracted increasing research attention. Various approaches for TSC have been explored, including statistical, signal processing, and machine or deep learning approaches in both time- and frequency domains. Despite intensive studies and rapid progress, important properties of many time series, such as shift equivariance and inversion invariance, remain underexplored in TSC. Moreover, existing methods face significant challenges in handling long-range dependencies, effectively utilizing both time and frequency representations, and maintaining computational efficiency. Many deep learning models, including convolutional neural networks (CNNs) and recurrent neural networks (RNNs), primarily focus on time-domain features, often overlooking frequency-based representations that could provide additional discriminatory power. Transformer-based approaches, while powerful, suffer from quadratic complexity in sequence length, making them less efficient for long time-series data. Meanwhile, state-space models (SSMs) like Mamba offer promising alternatives but have not been widely explored in TSC. These challenges, including the need for better handling of time series properties and computational efficiency, motivate our work in developing a unified, efficient, and multi-view TSC approach leveraging both spectral and temporal insights.

Equivariance refers to a property where transformations applied to the input lead to predictable and corresponding transformations in the output. In the case of shift equivariance for time series data, this means that when we apply a shift transformation to the input time series, the output of our model (such as detected features or patterns) undergoes a corresponding shift. Shift equivariance property is crucial for TSC as it allows pattern recognition regardless of exact temporal positions, offering benefits such as: 
\begin{itemize}
    \item Resilience to temporal misalignments between different instances of the same class, which is common in real-world data. 
    \item Creation of more consistent features across samples when dealing with MTS of varying lengths, such as when training and test examples have different durations. 
    \item Improved generalization while retaining information about temporal relationships, as the classifier can learn patterns that are consistent in their relative positions across different time shifts. 
\end{itemize}
These benefits collectively enhance the robustness and flexibility of TSC models, making them more adaptable to the diverse and often noisy nature of real-world time series data. However, existing approaches often struggle to fully exploit these advantages. While traditional statistical methods like dynamic time warping (DTW) excel at alignment, they struggle with complex temporal structures. Deep learning approaches each have their limitations: CNNs, though inherently shift-equivariant, struggle with long-range dependencies; Transformers offer flexible attention but at high computational cost; and SSMs like Mamba, while promising efficient sequence modeling, require specialized adaptations for TSC tasks. These limitations highlight the need for a TSC model that effectively balances efficiency, feature diversity, and sequence modeling capabilities.

Existing machine learning approaches for TSC often rely on convolution-based methods or neural networks to endow shift equivariance for the extracted features~\citep{lecun1989backpropagation}. Many models have been designed to exploit this property for visual learning tasks with 2D or 3D data~\citep{thomas2018tensor, fuchs2020se}. For TSC, several convolution-based methods~\citep{Dempster2020ROCKETEF, Franceschi2019UnsupervisedSR, IsmailFawaz2018deep} have implicitly utilized this property. However, they only employ temporal features by applying convolutions in the time domain.

In addition to time-based features, frequency-based features can help identify important patterns in the data's spectral composition. Disentangling time and frequency components has been shown to be critical for time series learning~\citep{zhang2022self, tslanet}. In particular, spectral features have been extracted using discrete Fourier transform (DFT)~\citep{tslanet} and discrete wavelet transform (DWT)~\citep{chaovalit2011discrete, ouyang2021adaptive, wavelet}. However, neither DFT nor DWT is shift equivariant - DFT lacks this property~\citep{oppenheim1999discrete}, while DWT's discrete nature and downsampling (decimation) step in its computation also prevent shift equivariance~\citep{mallat2008wavelet, percival2000wavelet}. Consequently, a small shift in the input signal can lead to significantly different coefficients or features for DFT and DWT. 

Continuous wavelet transforms (CWT) with real-valued mother wavelets possess shift equivariance~\citep{mallat2008wavelet}. CWT coefficients offer a localized time-frequency representation of the signal, with scale-dependent locality~\citep{mallat2008wavelet}. This means they capture more localized features at higher frequencies and broader features at lower frequencies, while still maintaining overall locality compared to global transforms like DFT. While CWT has been used for TSC~\citep{wang2020rollingCWT}, global patterns of MTS data have been insufficiently considered. MTS data often contain sensible global patterns, such as trends, seasonality, periodicity, cycles, and long-term dependencies. Lacking the ability to capture shift-equivariant global features can lead to a loss of discriminative information for TSC.

These considerations highlight that many existing approaches cannot fully exploit the shift equivariance of features or patterns for TSC. While some CNN or CWT-based methods can capture shift-equivariant features, they often lack the ability to effectively utilize spectral or global features. A clear need arises for an effective approach that can leverage shift-equivariant local and global features in both time and frequency domains to enhance TSC performance.
In this paper, we propose a novel approach for TSC that effectively leverages time shift-equivariant features and patterns from both time and frequency domains. The shift-equivariant spectral features are derived from the time-frequency representations of CWT with a real-valued mother wavelet. To remedy CWT's limitation regarding the locality of resulting features, we leverage convolutional kernels to extract shift-equivariant temporal features.

While convolutional kernels excel at capturing temporal dependencies between input features or inter-dependencies between channels, preserving shift equivariance, they typically have limited lengths, resulting in a limited receptive field that extracts temporally localized content \citep{bengio2013representation}. To enrich the expressiveness of CNN-based features, we adopt the kernel-based feature transformation ROCKET~\citep{Dempster2020ROCKETEF}. ROCKET uses random kernels with random lengths, potentially extracting a wide spectrum of temporal features, ranging from highly local to global ones.

However, many TSC tasks involve MTS data with characteristics or patterns spanning the entire length or a significant portion of the time series, suggesting that globally discriminative features or global interactions of features can be beneficial. While ROCKET may capture global features to a certain extent due to its use of kernels with random lengths, it tends to focus more on local features with varying degrees of locality.
To enhance the extraction of global features, we incorporate fully connected MLPs. These can capture temporally global patterns with their wide receptive field, although they may not be well-suited for identifying temporally local patterns or temporal dependencies due to their treatment of each input feature independently. Thus, we use MLPs to complement ROCKET's approach, strengthening the extraction of global features. Although hybrid approaches like ROCKET and MLP-based methods attempt to address local and global feature extraction, they do not inherently integrate spectral information or maintain shift equivariance across domains. Moreover, existing fusion mechanisms often lack adaptability, relying on fixed feature combinations rather than dynamic selection based on input characteristics. To address these limitations, we propose a novel feature fusion strategy that adaptively selects and integrates spectral and temporal features, leveraging their complementary strengths.

To avoid significantly increasing the size of extracted features, we propose a switch mechanism that selects between CNN-based (primarily local) features and MLP-based global patterns. This mechanism determines which type of temporal features - local or global - is more discriminative for a given input and integrates it with the CWT-domain features. Our experiments demonstrate that the switch mechanism effectively captures the most salient characteristics of the input data, whether they are local or global in nature.

The time-frequency representations from the CWT and temporal features from ROCKET transformation, or temporal features from global MLP, will be leveraged jointly to exploit the MTS characteristics, particularly shift equivariance, across domains. These different types of features provide various perspectives on the MTS data. We combine these perspectives with an approach known as multi-view learning, which has been shown to improve model performance and reliability, e.g.,  in image classification and clustering~\citep{peng2024fine}. This provides comprehensive, enriched multiview contexts to the subsequent inference, thus enhancing TSC performance.

In addition, we introduce inversion invariance, a new concept in TSC where time series' features or patterns are equally useful for classification when read in both forward and backward directions. This property is particularly relevant for data where the time direction is not inherently meaningful, such as ECG patterns, climate data, or rotational data with arbitrary start points. We posit that inversion invariance can generally enhance TSC performance for the following reasons:
\begin{itemize}
    \item Using inversion invariance for TSC can effectively double the amount of input data, since both forward and backward readings will be used to train the same model. This increase in training examples is a new form of  data augmentation, which leads to better generalization, reducing potential overfitting for TSC.
    \item Capturing inversion-invariant patterns may help enhance the model's robustness to noise or disturbances. MTS data are often noisy~\citep{kang2014detecting}, which can mask the underlying signals and affect the robustness of the algorithm. By identifying patterns meaningful in both directions, the TSC model may focus more on intrinsic patterns while being less affected by (potentially direction-specific) noise.
\end{itemize}

Inversion invariance can improve generalization and robustness for TSC, particularly when patterns are significant in both forward and reverse time directions. While Transformers and CNN-based architectures have been widely explored for time series modeling, their limitations in handling long sequences and computational overhead make them suboptimal for large-scale TSC tasks. SSMs like Mamba offer a promising alternative by efficiently modeling long-range dependencies with linear complexity. However, standard Mamba lacks explicit mechanisms for handling inversion invariance, a critical property in time-series data where temporal direction may be ambiguous (e.g., ECG signals). To overcome this, we introduce a novel scanning mechanism—tango scanning—that enhances inter-token relationships while preserving computational efficiency.
Our extensive experiments across various MTS datasets demonstrate the effectiveness of this tango scanning approach, supported by theoretical analysis using attention mechanisms.

To facilitate final classification, we employ Mamba~\citep{gu2023mamba}, a state-of-the-art (SOTA) model based on SSMs. Like RNNs, SSMs use state variables to represent the system's internal condition and its evolution over time~\citep{gu2021combining}. Mamba introduces selective state spaces, a mechanism that updates only a subset of state dimensions based on each input. This allows Mamba to focus on the most relevant information, efficiently process long sequences, and capture long-range dependencies. Mamba-based models have demonstrated competitive performance on various tasks, including language modeling~\citep{gu2023mamba,mamba2}, time series forecasting~\citep{atik2024timemachine}, deoxyribonucleic acid (DNA) sequence modeling~\citep{gu2023mamba}, tabular data learning~\citep{atik2024mambatab}, and audio generation~\citep{shams2024ssamba}. Unlike popular Transformers, which have quadratic time complexity in sequence length, Mamba achieves linear time complexity. This makes it more suitable for processing long sequences and scaling to larger datasets. By using Mamba, our TSC model is efficient in training and inference, with reduced computational costs and memory requirements compared to existing SOTA models.

We introduce a novel sequence scanning scheme, tango scanning, for inputting the original sequence and the reversed sequence into the same Mamba block and fusing the output. This scheme uses essentially the same memory footprint but demonstrates higher accuracy than vanilla Mamba scanning. Our tango scanning scheme differs from existing bi-directional Mamba implementations in the following ways: 1) Compared to~\citep{wang2024mamba, behrouz2024graph}, our approach uses one vanilla Mamba block for both directions, while theirs use two separate blocks. 2) Compared to BiMamba~\citep{schiff2024caduceus}, we use a single Mamba block without weight ties, and only one reversal operation, whereas BiMamba uses two blocks with partial weight ties and two reversal operations.
3) Compared to MambaDNA~\citep{schiff2024caduceus}, we use a single reversal operation and one Mamba block, while MambaDNA uses two ``reverse complement'' operations and two Mamba blocks with weight ties where the ``reverse complement'' operation is specific to DNA sequences to reflect the complementary nucleotides in the DNA base pairs.
Through extensive experiments, we demonstrate that our multi-view Mamba-based approach outperforms or matches the performance of existing SOTA models, typically with a small fraction of computational requirements and reduced memory usage.

In summary, the contributions of this paper include:
\begin{itemize}
    \item Building a novel multi-view approach for TSC: Our approach seamlessly integrates frequency- and time-domain features to exploit shift-equivariant patterns and provide complementary, discriminative contexts for TSC. It also employs a gating scheme to fuse spectral features with local or global time-domain features, effectively leveraging patterns characterizing the MTS classes.  
    \item Adapting the Mamba state-space model for sequence modeling and TSC: Our model leverages Mamba's linear complexity to efficiently capture long-term dependencies within MTS data, while incorporating well-motivated modifications tailored for TSC tasks. 
    \item Introduction of inversion invariance for TSC: This includes the new concept of inversion invariance for MTS classification. It also includes building an innovative Mamba-based ``tango scanning'' scheme to identify inversion invariant features or patterns. Our proposed tango scanning demonstrates improved effectiveness in modeling inter-token relationships in the sequence compared to vanilla Mamba block or bi-directional Mamba for TSC.     
    \item Extensive experimental validation of the proposed approach: Our model shows superior performance over various existing SOTA models over two sets of standard benchmarking datasets (10+20 datasets), achieving average accuracy improvements of 4.01-6.45\% and 7.93\% respectively over leading TSC models among 20 SOTA baselines. 
\end{itemize}
By addressing key limitations of existing approaches—including inefficient long-range dependency modeling, suboptimal fusion of spectral and temporal features, and lack of inversion-invariant learning—our approach provides a well-rounded, efficient solution for TSC. The proposed TSCMamba framework not only achieves superior performance over SOTA baselines but also lays the groundwork for future research in multi-view learning and efficient sequence modeling for time series data. These contributions are expected to advance real-world TSC applications in research and everyday life. The following sections will briefly review related works, present our approach in detail, and demonstrate extensive experimental results and ablation studies to conclude the paper.

\section{Related Works}
In this section, we provide a brief review of relevant methods for TSC in the literature, focusing on works using machine learning or deep learning. We group existing methods into 4 categories: traditional methods like DTW,  deep learning approaches using CNNs or RNNs, Transformer architectures, and methods based on state-space models.   
\subsection{Traditional Methods}
Traditional TSC methods include techniques like Dynamic Time Warping (DTW) \citep{Berndt1994UsingDT}, which measures the similarity between time series by aligning them in a non-linear way of dynamic programming. Tree-based methods like XGBoost \citep{Chen2016XGBoostAS} have also been applied to the TSC task.
\subsection{Deep Learning Approaches}
In recent years, deep learning approaches have become increasingly popular for TSC. Various MLP-based methods have been proposed, including DLinear by \citet{Zeng2022AreTE} and LightTS by \citet{Zhang2022LessIM}. DLinear constructs a simple model based on MLP, while LightTS uses light sampling-oriented MLP. These models are generally efficient in computations. Convolutional neural networks (CNNs) have been adapted for TSC, such as ROCKET \citep{Dempster2020ROCKETEF,minirocket} which uses random convolutional kernels for fast and accurate classification. ROCKET has inspired multiple extensions, including MiniROCKET~\citep{dempster2021minirocket} and MultiROCKET~\citep{tan2022multirocket}, which further optimize feature extraction efficiency and classification performance. These methods leverage thousands of randomly initialized convolutional kernels to capture diverse temporal patterns without requiring extensive training, making them effective for time series classification. Given its strength, ROCKET-based features have been widely integrated into hybrid models. Inspired by this, TSCMamba leverages ROCKET-derived representations to complement spectral features, enhancing discriminability in time series classification.
CNN has been also used by \citet{Franceschi2019UnsupervisedSR}  for learning representations of multivariate time series in an unsupervised way, which is then further leveraged for TSC. Besides CNNs, recurrent neural networks (RNNs) such as long-short-term memory (LSTM)  \citep{Hochreiter1997LongSM} 
and a variant, a gated recurrent unit (GRU), has also been adopted for TSC. Moreover, CNN and RNN have been combined to handle TSC effectively \citep{2018Modeling}.  
\subsection{Transformer-Based Methods}
Transformers by~\citet{NIPS2017_3f5ee243}, originally used for natural language processing, have been adapted for time series modeling. Reformer by \citet{kitaev2020reformer} introduces efficiency improvements to handle longer sequences. Numerous Transformer variants have been proposed to better model the unique characteristics of time series, such as handling non-stationarity with on-stationary Transformers by  \citet{Liu2022NonstationaryTR}, combining exponential smoothing with ETSformer in  \citet{woo2022etsformer}, and using decomposition and auto-correlation with Autoformer in \citet{wu2021autoformer}. Other variants include Pyraformer \citep{liu2021pyraformer}, which uses pyramidal attention to reduce complexity,  Flowformer \citep{wu2022flowformer}, which linearizes Transformers using conservation flows, and Informer  \citep{zhou2022fedformer}, which focuses on efficient long sequence forecasting by exploiting frequency-enhanced decomposition. 
Notably, TimesNet \citep{wu2022timesnet}  models the temporal 2D variations in time series data using a hierarchical structure of temporal blocks. By combining 2D convolutions, multi-head self-attention, and a novel positional encoding scheme, it can capture both local patterns and long-range dependencies in time series and obtain state-of-the-art performance. Recently, patch-based methods have gained attention in time series modeling, particularly in Transformer-based architectures. PatchTST~\citep{Yuqietal-2023-PatchTST} introduces a non-overlapping patching mechanism to improve long-range dependency modeling. Unlike sequence-to-sequence methods, PatchTST processes time series data in local patches, capturing finer-grained temporal structures.
Despite the impressive performance of Transformer-based models, \citet{Zeng2022AreTE} have shown that MLP-based models can be more effective in many scenarios.  

\subsection{State-Space Models}
Recently, a method called LSSL by \citet{gu2022efficiently} has been proposed for TSC with a structured SSM called S4. It employs a special parameterization called the Diagonal Plus Low-Rank form to represent the state transition matrix, enabling efficient computation over long sequences. Mamba~\cite{gu2023mamba} dynamically parameterizes state transitions based on input content, filtering out irrelevant information and selectively propagating key features. This mechanism has demonstrated strong results in sequence modeling, motivating its adaptation in TSCMamba to enhance time series classification performance. These innovations have made SSMs competitive alternatives to traditional RNN and Transformer-based models. Furthermore, recent works such as TimeMachine~\citep{atik2024timemachine} and S-Mamba~\citep{wang2024mamba} have demonstrated the potential of Mamba in time series forecasting, reinforcing the effectiveness of structured state-space models in capturing long-range dependencies.

\section{Methodology}
\begin{figure}[t]
    \centering
    \includegraphics[width=\linewidth]{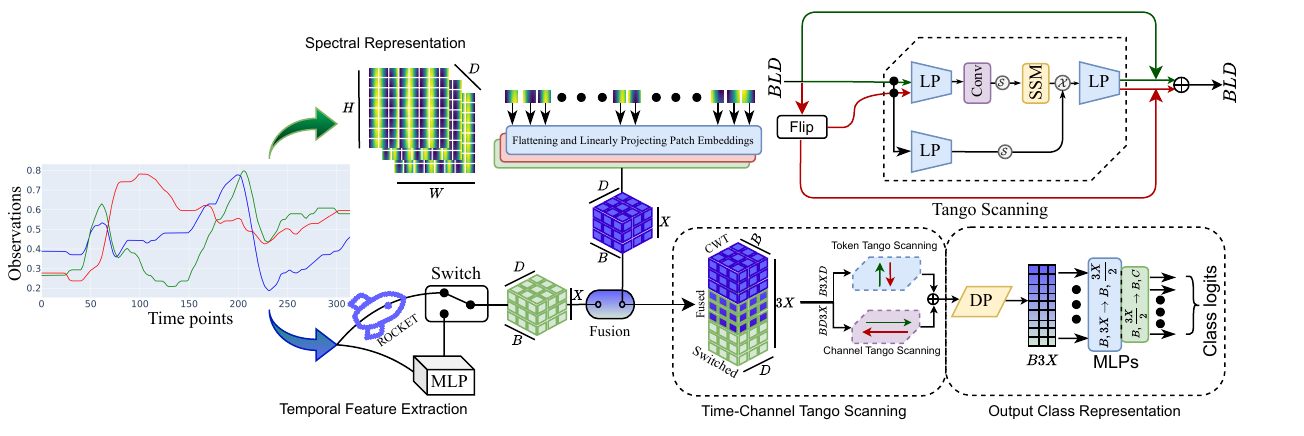}
    \caption{Schematic illustration of the proposed model TSCMamba. This diagram illustrates the architecture of our approach, featuring tango scanning. Here, DP refers to Depth-wise Pooling (DP) and LP refers to Linear Projection (LP). A switch gate selectively activates the utilization of either ROCKET or MLP-derived features. The MLP module, depicted in the bottom right, comprises two layers with an optional dropout mechanism interspersed for regularization. Here, $B$ = Batch, $L$ = Length, $D$ = Dimension.}
    \label{fig:method}
\end{figure}

This section describes our proposed method, TSCMamba, whose overall architecture is schematically illustrated in Fig. \ref{fig:method}. The architecture and data flow comprise the following key steps: First, we generate global temporal features, localized temporal features, and joint temporal-spectral features from the MTS data (Subsection~\ref{sub:spectral_representation} and~\ref{sub:temporal_feature}).
Second, these diverse views are fused to provide rich contexts for subsequent sequence modeling (Subsection~\ref{sec:fusion}).
A switch gate determines whether to fuse the transform-domain features with local features or global patterns in the temporal domain.
Third, both the pre-fusion and post-fusion features are fed into an inference engine consisting of two Mamba blocks. This captures long-term dependencies between these features (Subsection~\ref{sub:tango_scanning}).
Fourth, we introduce a tango scanning scheme for each Mamba block to exploit inversion-invariant features, followed by depth-wise pooling.
Fifth, final class decisions are made using an MLP to generate class logits (Subsection~\ref{sub:output_class}).
The following subsections detail each component and procedure of TSCMamba.

\subsection{Spectral Representation}
\label{sub:spectral_representation}
We have chosen Continuous Wavelet Transform (CWT) to represent raw signals in the spectral domain. CWT potentially surpasses space-time Fourier transform, fast Fourier transform (FFT), and digital wavelet transform (DWT) by providing superior time-frequency localization and multi-resolution analysis. CWT's adaptable wavelets enhance feature extraction and noise reduction while better-handling edge effects. This makes CWT particularly suitable for analyzing non-stationary signals. CWT's continuous and detailed representation may offer a significant advantage over the discrete nature of DWT. This renders CWT highly effective for precise time-frequency analysis. 

Among a variety of wavelets, the Morlet wavelet in Equation~\ref{eq:morl} is employed in this paper due to its capturing both amplitude and phase effectively:
\begin{equation}
    \psi(t) = (\pi^{-1/4}) (1 - \frac{t^2}{\sigma^2}) \exp(-\frac{t^2}{2\sigma^2}) \cos(2 \pi f t), 
      \label{eq:morl}
\end{equation}
where $\sigma$ is the scale parameter controlling the width of the wavelet, and  $f$ is the frequency parameter that controls the frequency of the cosine function. In this paper, we adopt $\sigma^2 = 1$ and $f=5/(2\pi)$ to balance computational cost and the expressiveness of the obtained wavelet features. However, keeping these parameters learnable may potentially benefit the classification accuracy. The smooth and symmetric shape of the Morlet wavelet minimizes distortions and edge effects, resulting in a clear and interpretable time-frequency representation. Using the wavelet function,  we obtain a 2-D representation of the size 
$L_1 \times L_1$ for each channel of an original MTS input sample of size $L$. In this paper, we adopt $L_1 = 64$ for computational efficiency and expressiveness of the obtained wavelet features. We summarize this CWT feature extraction process in S-Algorithm \ref{alg:cwt_conversion}. Since conversion from time signals to CWT representation is not learnable, we move this to the data pre-processing part, while regarding only the patch embedding module to be learnable. This helps our model to achieve lower FLOPs and faster training.

With the resultant CWT representation of size $D \times L_1 \times L_1$, we further perform patch embedding using a Conv2D layer (kernel size = stride = $p$, padding = 0), where $p = 8$ is patch size. Later with flattened patches, we utilize a feed-forward network (FFN) to obtain patches of size $D \times X$ for each MTS sample. The FFN consists of one fully connected layer with an input dimension $({\frac{L_1}{p}})^2$ and an output dimension $X$, as shown for the projected space in Figure~\ref{fig:method}. It is used to extract features within the CWT representation. 
For each batch of size $B$, the resultant tensor for representing CWT features is denoted by 
$\mathcal{W} \in {\mathcal{R}}^{B \times D \times X}$. 

\subsection{Temporal Feature Extraction}
\label{sub:temporal_feature}
To complement the frequency-domain features, we extract time-domain features. As previously discussed, 
different MTS datasets may have global or local features or patterns that discriminate between different classes. 
Capturing these features is essential for accurate classification. 
We leverage two different approaches to capture such features. 

\noindent {\bf{Extracting Local Features with Convolutional Kernels in Unsupervised Fashion.}}
Convolutional kernels usually have limited receptive fields, thus focusing on the extraction of local features. 
Since an MTS dataset may have local features at multiple temporal scales, it is sensible to capture local features within various widths of receptive fields. To this end, we employ the ROCKET approach \citep{Dempster2020ROCKETEF} to extract local features within various local neighborhoods for each channel in an unsupervised fashion. Here, it is to be noted that we only utilize the time domain to extract the kernel-based features, we do not utilize the class labels of the corresponding features. Therefore, our improvement in performance does not solely rely on the ROCKET method, rather it works as a performance booster in certain datasets.

ROCKET is a randomized algorithm that uses a set of randomized convolutional kernels to extract features from time series data. The method is suitable for capturing local features at various scales due to its randomized nature and the use of kernels with different sizes and strides. The procedure first randomly generates a set of convolutional kernels, each with a specific size and stride. Next, it convolves each kernel with the time series data to generate a feature map. The procedure is summarized in S-Algorithm \ref{alg:rocket}. ROCKET generates random convolutional kernels, including random length and dilation. It transforms the time series with two features per kernel. The global max pooling and the Proportion of Positive Values (PPV).  This fits one set of parameters for individual series.

We apply the ROCKET feature extraction method to each channel of length $L$ to form a feature vector of length $X$. We input our training data as a result, with the input tensor of size $B \times D \times L$, we obtain a tensor $\mathcal{V}_L \in {\mathcal{R}}^{B \times D \times X}$ that represents the local features. We utilized the sktime implementation of ROCKET to achieve this~\citep{franz_kiraly_2024_11460888,loning2019sktime}.

\noindent {\bf{Global Feature Extraction with MLP.}}
MLP has a receptive field covering the entire input, allowing the resulting feature vectors to capture the global characteristics of the MTS data. 
Independently for each channel of the input MTS of size $D \times L$, we utilize a one-layer MLP with linear activation to obtain a feature vector of size $D \times X$. 
Therefore, with the input tensor of size $B \times D \times L$, we obtain a tensor $\mathcal{V}_G \in {\mathcal{R}}^{B \times D \times X}$ representing the global features.

\subsection{Fusing Multi-View Representations}
\label{sec:fusion}
After obtaining the CWT-based spectral features using CWT and the temporal features at both local and global levels, 
we fuse these features to effectively exploit the complementary information in these multi-view representations. Through our empirical study, we observe that for many MTS data, either local features or global features in the temporal domain play a dominant role in discriminating between classes for TSC. This observation motivates us to fuse the spectral features with either global or local features in the temporal domain. Denote the temporal features by $\mathcal{V}$, which is either $\mathcal{V}_G$ or $\mathcal{V}_L$. Then, the fused feature map $\mathcal{V}_W$ will be calculated as follows: 
\begin{equation}
    \mathcal{V}_W = \mathcal{W} \otimes \mathcal{V},
\end{equation}
where $\otimes$ represents an element-wise operation. In this paper, this is either a multiplicative or additive operation such that 
\begin{equation}
    \{\mathcal{V}_W\}_{ijk} = \lambda  \mathcal{V}_{ijk} * (2 - \lambda)   \mathcal{W}_{ijk},\quad {\text{or,}} \quad  \{\mathcal{V}_W\}_{ijk} = \lambda  \mathcal{V}_{ijk} + (2 - \lambda)\mathcal{W}_{ijk}, 
\end{equation}
where $\lambda \ge 0$ is a learnable parameter that determines the balance between the spectral features and the temporal features, $1 \le i \le B$, $1 \le j \le D$, $1 \le k \le X$. We set the $\lambda$ as a learnable parameter while the initial value of $\lambda=1.0$. Therefore, the optimal value of $\lambda$  will be determined during the training process. The initial value of $\lambda=1.0$ ensures a balanced focus initially between temporal and spectral domain features.

After obtaining the fused  temporal-spectral features, we composite it with the tensors containing the multi-view features into a new tensor $\mathcal{U} = \mathcal{W} \| \mathcal{V}_W \| \mathcal{V} \in {\mathcal{R}}^{B \times D \times 3 X}$, where 
$\|$ is a concatenation operation. We use a switching mechanism to make the choice between  $\mathcal{V} = \mathcal{V}_G$ or $\mathcal{V} = \mathcal{V}_L$. This mechanism is implemented as a learnable binary mask that selects either the global or local temporal features during the training process. The final state of this switch will be determined by the optimization in the training process and tuned based on datasets for optimal performance.  

\subsection{Inferring with Time-Channel Tango Scanning}
\label{sub:tango_scanning}
With the integrated temporal-spectral contextual representations contained in tensor $\mathcal{U}$, which is processed by a layer normalization for training stability, we can now learn salient representations to capture important relationships between features, particularly long-term dependencies. To achieve this, we construct tokens by treating each feature vector in $\mathcal{U}$ 
along the time and channel dimensions as a separate token. Subsequently, we leverage Mamba, a type of SSM, for modeling the token sequences. Mamba is designed for capturing discriminative contents by selectively scanning the token sequences. This selective scanning ability allows the model to focus on the most informative parts of the sequence while ignoring less relevant information. By doing so, Mamba can effectively capture long-term dependencies and identify salient features that are most useful for classification. 

Compared to other SSMs, Mamba has the advantage of being computationally efficient and able to handle long sequences. It achieves this by using a sparse attention mechanism that reduces the complexity of token-to-token interactions. This makes Mamba particularly well-suited for processing time series data, where the sequences can be lengthy and contain complex temporal dependencies. 

\noindent {\bf{Vanilla Mamba Block:}} Inside a Mamba block, two fully-connected layers in two branches calculate linear projections. The output of the linear mapping in the first branch passes through a 1D causal convolution and SiLU activation ${\mathcal{S}}(\cdot)$~\citep{elfwing2018sigmoid}, then a structured SSM. The continuous-time SSM maps an input function or sequence $u(t)$ to output $z(t)$ through a latent state $h(t)$: 
\begin{equation}
    \label{eq:CT-SSM}
    dh(t)/dt = A \ h(t) + B \  u(t), \quad z(t) = C \ h(t), 
\end{equation}
where $h(t)$ is $N$-dimensional, with $N$ also known as a {\sl{state expansion factor}}, $u(t)$ is $D$-dimensional, with $D$ being the {\sl{dimension factor}} for an input token, $z(t)$ is an output of dimension $D$, and $A$, $B$, and $C$ are coefficient matrices of proper sizes. 
This dynamic system induces a discrete SSM governing state evolution and outputs given the input token sequence through time sampling at $\{ k \Delta \}$ with a $\Delta$ time interval. This discrete SSM is 
\begin{equation}
    \label{eq:DT-SSM}
    h_k = \bar{A} \ h_{k-1} + \bar{B} \ u_{k}, \quad z_k = C \ h_{k}, 
\end{equation}
where $h_k$, $u_k$, and $z_k$ are respectively samples of $h(t)$, $u(t)$, and $z(t)$ at time $k \Delta$, 
\begin{equation}
    \bar{A} = \exp(\Delta A), \quad 
    \bar{B} = (\Delta A)^{-1} (\exp(\Delta A) - I) \Delta B.    
\end{equation}
For SSMs, diagonal $A$ is often used. Mamba makes $B$, $C$, and $\Delta$ linear time-varying functions dependent on the input. In particular, for a token $u$,  
%\begin{equation}
%\begin{split}
$B,C  \leftarrow Linear_N(u)$,  and $\Delta  \leftarrow softplus(parameter + Linear_D(Linear_1 (u)))$, 
%\end{split}
%\end{equation}
where $Linear_p (u)$ is a linear projection to a $p$-dimensional space, and $softplus$ activation function. 
Furthermore, Mamba also has an option to expand the model dimension factor $D$ by a controllable dimension expansion factor $E$.  
Such coefficient matrices enable context and input selectivity properties \citep{gu2023mamba} to selectively propagate or forget information along the input token sequence based on the current token. Denote the discretization operation by $\Delta = \tau_{\Delta} (parameter + s_{\Delta})$,  where $\tau_{\Delta}$ and $s_{\Delta}$ are both functions of the input. For the special case of univariate sequences, the selectivity property has been mathematically proved~\citep{gu2023mamba}, as shown in the following:

\begin{theorem} 
(Gu and Dao 2023) When $N = 1, A = -1, B = 1$, $s_{\Delta} = Linear(x)$, and $\tau_{\Delta} = softplus$, 
then the selective SSM recurrence takes the form of 
\begin{equation}
    h_k = (1-g_k) \ h_{k-1} + g_k \ u_k, \quad {\text{and}} \quad g_k = \sigma(Linear(u_k)),
\end{equation} 
where $g_k$ is the gate. 
\end{theorem}
This theorem states that the hidden state is a convex combination of the current input token and the previous hidden state, with the combination coefficient controlled by the current input token. Moreover, it is pointed out that the parameter $g_k$ is responsible for selecting the input contents $u_k$ from the sequence, plays a role similar to a gating mechanism in the RNN model, thus connecting the selective SSM to the traditional RNN. 

After obtaining the SSM output, it is multiplicatively modulated with the output from the second branch before another fully connected projection.  The second branch in the Mamba block simply consists of a linear mapping followed by a SiLU. 

\noindent {\bf{Tango Scanning:}} 
\label{sec:tango_scaning}The selectivity ability of Mamba depends on the ordering of the tokens in the sequence because the hidden state at time $n$ is constructed causally from history tokens as determined by the ordering of the tokens. If the history tokens do not contain informational contexts, Mamba may provide less effective predicted output. To alleviate this potential limitation of causal scanning, we construct a dedicated module to extend a vanilla Mamba block, as shown in Figure \ref{fig:method}.  Each module comprises one vanilla Mamba block. On the input side, the module accepts a sequence in a forward fashion as input and then inverts the sequence to accept it as input again. At the output side, the output of the vanilla Mamba block with forward sequence and that with the inverted sequence are added element-wise. The operations are represented as follows. Denote an input token sequence by $v = [v_1, \cdots, v_M]$, where $v_i \in {\mathcal{R}}^D$, and $v \in {\mathcal{R}}^{D \times M}$ is the matrix representation of the token sequence with $M$ being the sequence length.  We will first get a reverse-flipped sequence  $v^{(r)}$  by inverting the ordering of the elements in $v$. Tango scanning performs the following operations to obtain the output sequence $s^{(o)}$: 
\begin{align}
   &  v^{(r)} = Reverse( v ) = [v_M, v_{M-1}, \cdots, v_1],\\ 
   & a = Mamba(v), \quad  a^{(r)} = Mamba(v^{(r)}), \\
 \label{eq_fusion}  & s^{(o)} = v \oplus a \oplus v^{(r)} \oplus a^{(r)} , 
\end{align}
where $Reverse(\cdot)$ denotes the flipping operation of a sequence, $Mamba(\cdot)$ denotes a vanilla Mamba block, and $\oplus$ denotes element-wise addition. The last equation {\ref{eq_fusion}} represents the element-wise addition for information fusion. Notably, the same Mamba block is used for the forward sequence $v$ and the reverse-flipped sequence  $v^{(r)}$.  By doing so, the SSM in this block will be trained to update the hidden state variable more effectively than using simply the forward scanning of the vanilla Mamba.  Because of the sharing of one Mamba block (and thus one SSM) with two sequences that are flips of each other, we regard it as a dancer's one body with two concerted legs and hence call it tango scanning.  

Unlike the bi-directional Mamba block in \cite{behrouz2024graph,schiff2024caduceus} that uses two separate SSMs with one for forward direction and the other for backward direction, our tango scanning block uses only a single Mamba block. Importantly,  for the inverted sequence in our tango scanning module, the output from the Mamba block is not re-inverted back to the original order before performing element-wise addition. In other words, a tango scanning module only involves sequence inversion once. On the contrary,  the bi-directional Mamba block  \citep{schiff2024caduceus} needs to re-invert the output from the vanilla Mamba block. 
Empirically, we will demonstrate that our tango scanning can effectively update the hidden state variable while maintaining essentially the same memory footprint as the vanilla Mamba block.                 

\noindent {\bf{Performing Tango Scanning in Time and Channel Dimensions:}} The MTS data have significant patterns, correlation structures, and temporal long-term dependencies. To model the relationship in the temporal dimension, we perform tango scanning temporally for every channel. 
The processed embedded representation with tensor size $B \times 3X \times D$ is transformed using tango scanning. Specifically,  with each  $D$-dimensional feature point across all channels regarded as a token,  we have a token sequence with dimension factor $D$ and length $3X$ as input to the Mamba block in the tango scanning module. This yields an output tensor of size   $B \times 3X \times D$.  
That is, by denoting $u^{(t)} = [u^{(t)}_{1}, \cdots, u^{(t)}_{3X}]^T \in {\mathcal{R}^{3X \times D}}$ as the token sequence formed along the time direction for a time series (in the batch), we have  
\begin{equation}
\label{eq_tango_time}
    s^{(t)} = Tango\_Scanning(u^{(t)}),
\end{equation}
where $s^{(t)} = [s^{(t)}_1, \cdots, s^{(t)}_{3X}]^T \in {\mathcal{R}^{3X \times D}}$. By leveraging Mamba, we will extract salient features and context cues from the input token sequence. Particularly, the output sequence $s_t$ captures the between-time-point interactions along the temporal direction. 

Because the MTS data often have significant correlations along the channel dimension, we will also model relationships across channels. To this end, we first form our tensor to have size   $B  \times D \times 3X$ and then we transform it using our tango scanning. Specifically, the whole univariate sequence of each channel is used as a token with a dimension factor $3X$ for the Mamba block in the tango scanning module. Thus, we form a token sequence of length $D$,  with each token having dimension $3X$. This token sequence will be input to our tango scanning module, yielding an output tensor of size  $B  \times D \times 3X$. 
That is, by denoting $u^{(c)} = [u^{(c)}_1, \cdots, u^{(c)}_D]^T \in {\mathcal{R}^{D \times 3X}}$ as the token sequence formed along the channel dimension, we have  
\begin{equation}
\label{eq_tango_channel}
    s^{(c)} = Tango\_Scanning( u^{(c)} ),
\end{equation}
where $s^{(c)} = [s^{(c)}_1, \cdots, s^{(c)}_D]^T \in {\mathcal{R}^{D \times 3X}}$. Note that the tango scanning module used in Eq. (\ref{eq_tango_channel}) is different from the one used in Eq. (\ref{eq_tango_time}) and utilizes a separate Mamba module. 
The output sequence $s_c$ captures the between-channel interactions along the temporal direction. It is critical to account for the inter-relationships across channels when the MTS data have many channels. 

After obtaining the outputs from the time-wise scanning $u_t$ and the channel-wise scanning $u_c$, we will perform another fusion at the Mamba-transformed sequence level:  
\begin{equation}
{z} = (s^{(t)})^T \oplus s^{(c)},
\end{equation}
where $\oplus$ denotes element-wise addition of matrices,  and $(s^{(t)})^T$ is the transpose of $s^{(t)}$. The resultant fused sequence is of size $D\times 3X$. 

\subsection{Output Class Representation}
\label{sub:output_class}
The fused tensor of size $B \times D  \times 3X$ will be used to distill class information (class logits). First, we perform depth-wise pooling (DP) (Figure~\ref{fig:method}) to aggregate information across channels. Specifically, given a fused sequence $z \in {\mathcal{R}^{D \times 3X}}$,  we have
\begin{equation}
    \bar{z} = DP(z),
\end{equation}
where $DP(\cdot)$ denotes the depth-wise pooling and the output $\bar{z} \in {\mathcal{R}^{3X}}$. DP can be either average pooling or max pooling. We regard these two pooling operations as two possible values of DP.  Given a dataset, the specific pooling will be determined in the training stage. 

Subsequently, we will employ an FFN of two layers with an optional dropout mechanism interspersed for regularization: 
\begin{equation}
    \bar{z}^{(1)} = MLP(\bar{z}), \quad 
    \bar{z}^{(2)} = MLP(\bar{z}^{(1)}), 
\end{equation}
where $\bar{z}$  is projected into vectors $\bar{z}^{(1)} \in {\mathcal{R}^{3X/2}} $ and $\bar{z}^{(2)} \in {\mathcal{R}^{C}}$.  The class labels will be determined based on  $\bar{z}^{(2)}$.  To train the proposed network, which we call TSCMamba, we employ a cross-entropy loss (CE) on the output of the second layer of the FFN,  $\bar{z}^{(2)}$.  

\subsection{Explanation of Our Scanning Mechanism}
\label{sub:explain_tango}
In this section, we provide a more comprehensive explanation of our scanning protocol. 
Although Mamba~\citep{gu2023mamba} was developed as a distinct architecture, its outstanding performance across various domains can be interpreted through the lens of the attention mechanism~\citep{NIPS2017_3f5ee243}.
Building on this theoretical framework and drawing inspiration from~\citet{ali2024hidden}, we analyze the theoretical foundations of the tango scanning mechanism. Consider a sequence of length $L$ along with its corresponding matrices $\bar{A}, \bar{B}, C$ from Equation~\ref{eq:DT-SSM}. With per-channel time-variant system matrices $\bar{A}_1, \cdots, \bar{A}_L$, $\bar{B}_1, \cdots, \bar{B}_L$, and $C_1, \cdots, C_L$, each channel within the selective state-space layers can be processed independently. For simplicity, let us temporarily assume that the input sequence $u$ consists of a single channel. Given the initial condition $h_{0}=0$, unrolling Eq.~\ref{eq:DT-SSM} yields:

\begin{equation} \label{eq:unrolling1}
 h_1 =  \bar{B}_1  u_1 ,\quad z_1 = C_1 \bar{B}_1 u_{1},\quad  h_2 = \bar{A}_2 \bar{B}_1 u_{1} + \bar{B}_2 x_{2}  ,\quad z_2 = C_2 \bar{A}_2 \bar{B}_1 u_{1} + C_2 \bar{B}_2 u_{2} 
\end{equation}

\begin{equation} \label{eq:unrolling3}
 h_t = \sum_{j=1}^t \big{(} \Pi_{k=j+1}^t \bar{A}_k \big{)} \bar{B}_{j}u_j, %
 \quad %
 z_t = C_t\sum_{j=1}^t \big{(} \Pi_{k=j+1}^t \bar{A}_k \big{)} \bar{B}_{j}u_j
\end{equation}

\begin{equation}\label{eq:MAMbaASmatmul}
z = \tilde{\alpha} u, \quad
\begin{bmatrix}
z_1 \\
z_2\\ 
\vdots \\
z_L \\
\end{bmatrix} 
=
\begin{bmatrix}
    C_1 \bar{B}_1 & 0 & \cdots & 0 \\
    C_2 \bar{A}_2 \bar{B}_1 & C_2 \bar{B}_2 & \cdots & 0 \\
    \vdots & \vdots & \ddots & 0 \\
    C_L \Pi_{k=2}^L \bar{A}_k \bar{B}_{1} \quad & C_L \Pi_{k=3}^L \bar{A}_k \bar{B}_{2} \quad & \cdots \quad & C_L \bar{B}_L
\end{bmatrix}
\begin{bmatrix}
u_1 \\
u_2\\ 
\vdots \\
u_L \\
\end{bmatrix}
\end{equation}

The matrix $\tilde{\alpha} \in \mathbb{R}^{L \times L}$ is a function of the input and other relevant parameters. The element at row $i$ and column $j$ captures how $u_j$ influences $z_i$, and is computed according to:
\begin{equation}\label{eq:attnPerlocation}
    \tilde{\alpha}_{i,j} = C_i \Big{(}\Pi_{k=j+1}^i \bar{A}_k \Big{)} \bar{B}_j
\end{equation}
Equations~\ref{eq:MAMbaASmatmul} and \ref{eq:attnPerlocation} establish the connection between $\tilde{\alpha}$ and the standard attention matrix, revealing that S6 operates as a variant of causal self-attention. This form of causality is beneficial for next-token prediction tasks, such as in the natural language processing domain for text generation or in time series forecasting tasks, where the current token must not have access to future tokens. However, in our time series classification tasks, this causality constraint is unnecessary since the entire sequence is available during processing. 
\begin{figure}[H]
    \centering
    \begin{tikzpicture}[
        node distance=1.5cm and 1.2cm, 
        every node/.style={circle, draw, minimum size=1cm, inner sep=0pt},
        every path/.style={->, >=stealth, thick}
    ]

    % Define nodes
    \node (u1) {$u_1$};
    \node[right=of u1] (u2) {$u_2$};
    \node[right=of u2] (u3) {$u_3$};
    \node[right=of u3] (u4) {$u_4$};
    \node[right=2.5cm of u4] (ul) {$u_L$};

    % Self-loops on all nodes
    \path (u1) edge [loop below] (u1);
    \path (u2) edge [loop below] (u2);
    \path (u3) edge [loop below] (u3);
    \path (u4) edge [loop below] (u4);
    \path (ul) edge [loop below] (ul);

    % % Reversed edges
    \path (u2) edge [bend right=30] (u1);
    \path (u3) edge [bend right=30] (u2);
    \path (u3) edge [bend right=30] (u1);
    \path (u4) edge [bend right=30] (u3);
    \path (u4) edge [bend right=30] (u2);
    \path (u4) edge [bend right=30] (u1);
    \path (ul) edge [bend right=30] (u4);
    \path (ul) edge [bend right=30] (u3);
    \path (ul) edge [bend right=30] (u2);
    \path (ul) edge [bend right=30] (u1);
       
    % % Dashed line (more visible)
    \draw[dotted, thick, -] (u4) -- (ul);

    \end{tikzpicture}

    \caption{Regular forward scanning process. The arrow from node $a$ to node $b$ means $a$ depends on $b$.}
    \label{fig:regular_mamba_scan}
\end{figure}

\begin{figure}[H]
    \centering
    \begin{tikzpicture}[
        node distance=1.5cm and 1.2cm, 
        every node/.style={circle, draw, minimum size=1cm, inner sep=0pt},
        every path/.style={->, >=stealth, thick}
    ]

    % Define nodes
    \node (u1) {$u_1$};
    \node[right=of u1] (u2) {$u_2$};
    \node[right=of u2] (u3) {$u_3$};
    \node[right=of u3] (u4) {$u_4$};
    \node[right=2.5cm of u4] (ul) {$u_L$};

    % Self-loops on all nodes
    \path (u1) edge [loop above] (u1);
    \path (u2) edge [loop above] (u2);
    \path (u3) edge [loop above] (u3);
    \path (u4) edge [loop above] (u4);
    \path (ul) edge [loop above] (ul);

    % % Reversed edges
    \path (u1) edge [bend right=30] (u2);
    \path (u1) edge [bend right=30] (u3);
    \path (u1) edge [bend right=30] (u4);
    \path (u1) edge [bend right=30] (ul);
    \path (u2) edge [bend right=30] (u3);
    \path (u2) edge [bend right=30] (u4);
    \path (u2) edge [bend right=30] (ul);
    \path (u3) edge [bend right=30] (u4);
    \path (u3) edge [bend right=30] (ul);

    % % Dashed line (more visible)
    \draw[dotted, thick, -] (u4) -- (ul);

    \end{tikzpicture}

    \caption{Flipped sequence scanning process.}
    \label{fig:flipped_mamba_scan}
\end{figure}

\begin{figure}[H]
    \centering
    \begin{tikzpicture}[
        node distance=1.5cm and 1.2cm, 
        every node/.style={circle, draw, minimum size=1cm, inner sep=0pt},
        every path/.style={->, >=stealth, thick}
    ]

    % Define nodes
    \node (u1) {$u_1$};
    \node[right=of u1] (u2) {$u_2$};
    \node[right=of u2] (u3) {$u_3$};
    \node[right=of u3] (u4) {$u_4$};
    \node[right=2.5cm of u4] (ul) {$u_L$};

    % Self-loops on all nodes
    \path (u1) edge [loop above] (u1);
    \path (u1) edge [loop below] (u1);
    \path (u2) edge [loop above] (u2);
    \path (u2) edge [loop below] (u2);
    \path (u3) edge [loop above] (u3);
    \path (u3) edge [loop below] (u3);
    \path (u4) edge [loop above] (u4);
    \path (u4) edge [loop below] (u4);
    \path (ul) edge [loop above] (ul);
    \path (ul) edge [loop below] (ul);
    
    \path (u2) edge [bend right=30] (u1);
    \path (u3) edge [bend right=30] (u2);
    \path (u3) edge [bend right=30] (u1);
    \path (u4) edge [bend right=30] (u3);
    \path (u4) edge [bend right=30] (u2);
    \path (u4) edge [bend right=30] (u1);
    \path (ul) edge [bend right=30] (u4);
    \path (ul) edge [bend right=30] (u3);
    \path (ul) edge [bend right=30] (u2);
    \path (ul) edge [bend right=30] (u1);
    % % Reversed edges
    \path (u1) edge [bend right=30] (u2);
    \path (u1) edge [bend right=30] (u3);
    \path (u1) edge [bend right=30] (u4);
    \path (u1) edge [bend right=30] (ul);
    \path (u2) edge [bend right=30] (u3);
    \path (u2) edge [bend right=30] (u4);
    \path (u2) edge [bend right=30] (ul);
    \path (u3) edge [bend right=30] (u4);
    \path (u3) edge [bend right=30] (ul);

    % % Dashed line (more visible)
    \draw[dotted, thick, -] (u4) -- (ul);

    \end{tikzpicture}

    \caption{Conceptual illustration of the tango scanning mechanism for enhancing inversion invariance in TSC.}
    \label{fig:combined_scan}
\end{figure}
\begin{figure}[H]
    \centering
    \subfloat[Regular scanning attention.]{{\includegraphics[width=0.30\linewidth]{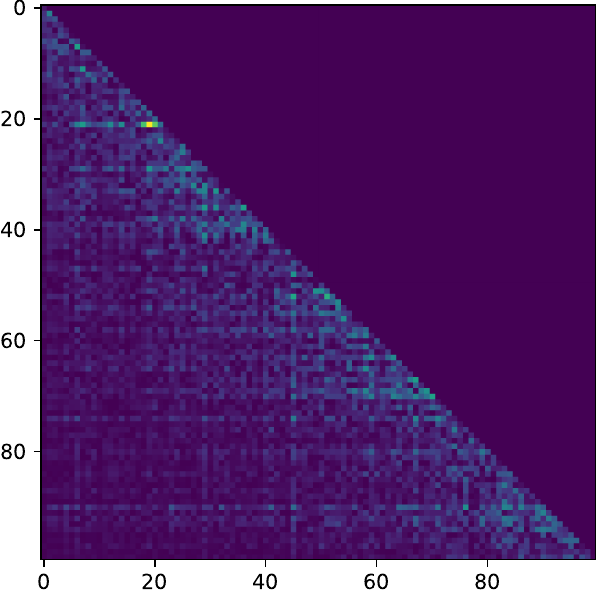} }}%
    \subfloat[Flipped scanning attention.]{{\includegraphics[width=0.30\linewidth]{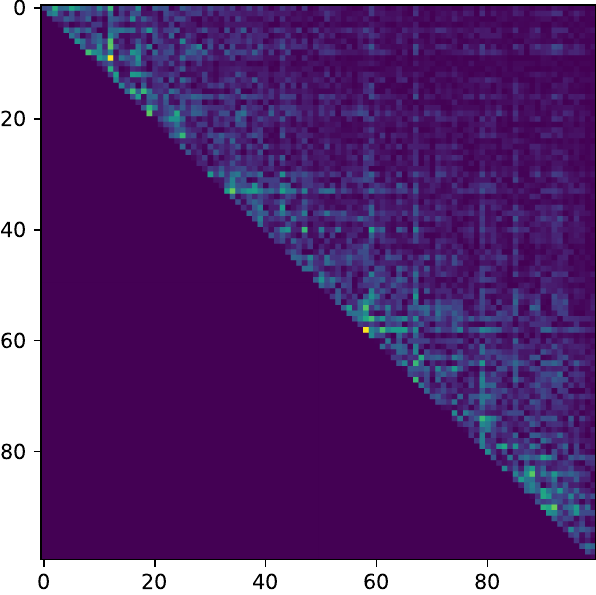} }}%
    \subfloat[Tango scanning attention.]{{\includegraphics[width=0.30\linewidth]{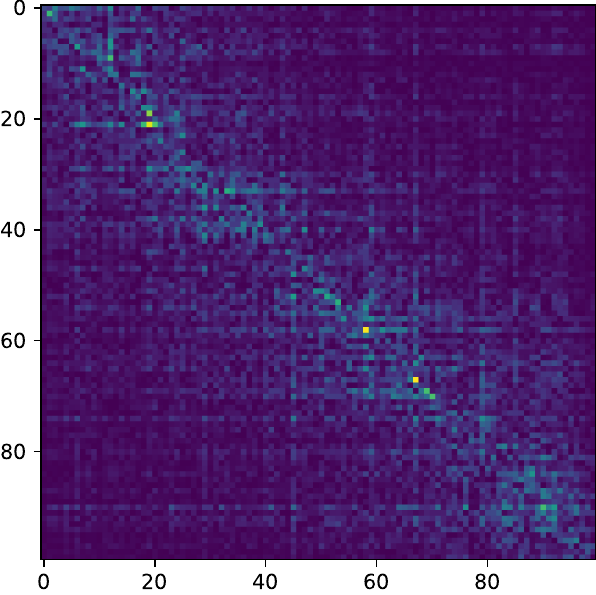} }}%
    \caption{Attention matrix visualization for different scanning protocols, illustrating the attended regions. Tango scanning achieves the maximum possible pairwise attention coverage with one Mamba.}
    \label{fig:attn_matrix_vis}
\end{figure}
Let $u$ be a time series of length $L$. When using only the regular Mamba, a causal attention mask~\citep{ali2024hidden} is applied (Figure~\ref{fig:regular_mamba_scan}), restricting the current token $u_i$ from attending to any future token $u_{i+1}$. To capture bidirectional dependencies, we introduce a flip-based scanning approach (Figure~\ref{fig:flipped_mamba_scan}), where each token $u_i$ attends to future tokens $u_{i+j}$ with $j \ge 1$. By combining both scanning directions with shared parameters (Figure~\ref{fig:combined_scan}), we achieve full pairwise attention using a single Mamba block without introducing additional parameters. Although this approach may introduce additional calculations for self-terms, it ensures maximum possible attention coverage, leaving no token unattended. We illustrate this strategy for a sequence length using token tango scanning, which can also be extended to channel tango scanning, where each channel is treated as a sequence length and each length as a channel. The practical attention matrix visualization in Figure~\ref{fig:attn_matrix_vis} demonstrates that tango scanning successfully captures all possible token pair interactions.

\section{Experiments and Result Analysis}
\begin{table}[H]
  \caption{Classification Accuracy (\%) for Various Models. The . symbol in Transformer models denotes the specific type of $\ast$former used. The best average result and rank is in \textbf{bold} and second best is \underline{underlined}. The ranks are computed by averaging dataset-wise rankings of classifiers (lower is better).}\label{tab:full_classification_results}
  \vskip 0.05in
  \centering
  \begin{threeparttable}
  \begin{small}
  \renewcommand{\multirowsetup}{\centering}
  \setlength{\tabcolsep}{1pt}
  \resizebox{\linewidth}{!}{
  \begin{tabular}{l|ccccccccccccccccccccccc}
    \toprule
    \multirow{3}{*}{Dataset} & \multicolumn{21}{c}{Methods} \\
    & TSCMamba & DTW  & XGBoost& Rocket & LSTM & LSTNet & LSSL & TCN & Trans & Re. & In. & Pyra & Auto. & Station. &  FED. & \update{ETS.} &Flow. & DLinear & LightTS. & TimesNet & TSLANet \\
	& (Ours) & \scalebox{1}{\citeyearpar{Berndt1994UsingDT}} & \scalebox{1}{\citeyearpar{Chen2016XGBoostAS}} &  \scalebox{1}{\citeyearpar{Dempster2020ROCKETEF}} & \scalebox{1}{\citeyearpar{Hochreiter1997LongSM}} & 
	\scalebox{1}{\citeyearpar{2018Modeling}} & 
	\scalebox{1}{\citeyearpar{gu2022efficiently}} & 
	\scalebox{1}{\citeyearpar{Franceschi2019UnsupervisedSR}} & \scalebox{1}{\citeyearpar{NIPS2017_3f5ee243}} & 
	\scalebox{1}{\citeyearpar{kitaev2020reformer}} & \scalebox{1}{\citeyearpar{haoyietal-informer-2021}} & \scalebox{1}{\citeyearpar{liu2021pyraformer}} &
	\scalebox{1}{\citeyearpar{wu2021autoformer}} & 
	\scalebox{1}{\citeyearpar{Liu2022NonstationaryTR}} &
	\scalebox{1}{\citeyearpar{zhou2022fedformer}} & \scalebox{1}{\citeyearpar{woo2022etsformer}} & \scalebox{1}{\citeyearpar{wu2022flowformer}} & 
	\scalebox{1}{\citeyearpar{Zeng2022AreTE}} & \scalebox{1}{\citeyearpar{Zhang2022LessIM}} & \scalebox{1}{\citeyearpar{wu2022timesnet}} &
    \scalebox{1}{\citeyearpar{tslanet}}\\
    \midrule
	\scalebox{1}{EC}& \scalebox{1}{62.0} & \scalebox{1}{32.3} & \scalebox{1}{43.7} & \scalebox{1}{45.2} & \scalebox{1}{32.3} & \scalebox{1}{39.9} & \scalebox{1}{31.1}&  \scalebox{1}{28.9} & \scalebox{1}{32.7} &\scalebox{1}{31.9} &\scalebox{1}{31.6}   &\scalebox{1}{30.8} &\scalebox{1}{31.6} &\scalebox{1}{32.7} &\scalebox{1}{31.2} & \scalebox{1}{28.1} & \scalebox{1}{33.8} & \scalebox{1}{32.6} &\scalebox{1}{29.7} & \scalebox{1}{35.7} &\scalebox{1}{30.4}\\
	\scalebox{1}{FD}&\scalebox{1}{69.4}   & \scalebox{1}{52.9} & \scalebox{1}{63.3} & \scalebox{1}{64.7} & \scalebox{1}{57.7} & \scalebox{1}{65.7} & \scalebox{1}{66.7} & \scalebox{1}{52.8} & \scalebox{1}{67.3} & \scalebox{1}{68.6} &\scalebox{1}{67.0} &\scalebox{1}{65.7} &\scalebox{1}{68.4} &\scalebox{1}{68.0} &\scalebox{1}{66.0} & \scalebox{1}{66.3} & \scalebox{1}{67.6} &\scalebox{1}{68.0} &\scalebox{1}{67.5} & \scalebox{1}{68.6}  & \scalebox{1}{66.8}  \\
	\scalebox{1}{HW}  & \scalebox{1}{53.3} & \scalebox{1}{28.6} & \scalebox{1}{15.8} & \scalebox{1}{58.8} & \scalebox{1}{15.2} & \scalebox{1}{25.8} & \scalebox{1}{24.6} & \scalebox{1}{53.3} & \scalebox{1}{32.0} & \scalebox{1}{27.4} &\scalebox{1}{32.8} &\scalebox{1}{29.4} &\scalebox{1}{36.7} &\scalebox{1}{31.6} &\scalebox{1}{28.0} &  \scalebox{1}{32.5} & \scalebox{1}{33.8} & \scalebox{1}{27.0} &\scalebox{1}{26.1} & \scalebox{1}{32.1}  & \scalebox{1}{57.9}\\
	\scalebox{1}{HB} & \scalebox{1}{76.6} & \scalebox{1}{71.7}  & \scalebox{1}{73.2} & \scalebox{1}{75.6} & \scalebox{1}{72.2} & \scalebox{1}{77.1} & \scalebox{1}{72.7}& \scalebox{1}{75.6} & \scalebox{1}{76.1} & \scalebox{1}{77.1} &\scalebox{1}{80.5} &\scalebox{1}{75.6} &\scalebox{1}{74.6} &\scalebox{1}{73.7} &\scalebox{1}{73.7} &  \scalebox{1}{71.2} & \scalebox{1}{77.6} & \scalebox{1}{75.1} &\scalebox{1}{75.1} & \scalebox{1}{78.0} & \scalebox{1}{77.6}\\
	\scalebox{1}{JV}  & \scalebox{1}{97.0}  & \scalebox{1}{94.9} & \scalebox{1}{86.5} & \scalebox{1}{96.2} & \scalebox{1}{79.7} & \scalebox{1}{98.1} & \scalebox{1}{98.4} & \scalebox{1}{98.9} & \scalebox{1}{98.7} & \scalebox{1}{97.8} &\scalebox{1}{98.9} &\scalebox{1}{98.4} &\scalebox{1}{96.2} &\scalebox{1}{99.2} &\scalebox{1}{98.4} & \scalebox{1}{95.9} &  \scalebox{1}{98.9} & \scalebox{1}{96.2} &\scalebox{1}{96.2} & \scalebox{1}{98.4}  & \scalebox{1}{99.2}\\
	\scalebox{1}{PS} &\scalebox{1}{90.2}  & \scalebox{1}{71.1} & \scalebox{1}{98.3} & \scalebox{1}{75.1} & \scalebox{1}{39.9} & \scalebox{1}{86.7} & \scalebox{1}{86.1}& \scalebox{1}{68.8} & \scalebox{1}{82.1} & \scalebox{1}{82.7} &\scalebox{1}{81.5} &\scalebox{1}{83.2} &\scalebox{1}{82.7} &\scalebox{1}{87.3} &\scalebox{1}{80.9} & \scalebox{1}{86.0} &  \scalebox{1}{83.8} & \scalebox{1}{75.1} &\scalebox{1}{88.4} & \scalebox{1}{89.6} &  \scalebox{1}{83.8}\\
	\scalebox{1}{SCP1}  & \scalebox{1}{92.5} & \scalebox{1}{77.7}  & \scalebox{1}{84.6} & \scalebox{1}{90.8} & \scalebox{1}{68.9} & \scalebox{1}{84.0} & \scalebox{1}{90.8} & \scalebox{1}{84.6} & \scalebox{1}{92.2} & \scalebox{1}{90.4} &\scalebox{1}{90.1} &\scalebox{1}{88.1} &\scalebox{1}{84.0} &\scalebox{1}{89.4} &\scalebox{1}{88.7} & \scalebox{1}{89.6} & \scalebox{1}{92.5} & \scalebox{1}{87.3} &\scalebox{1}{89.8} & \scalebox{1}{91.8}  & \scalebox{1}{91.8}   \\
    \scalebox{1}{SCP2} & \scalebox{1}{66.7} & \scalebox{1}{53.9} & \scalebox{1}{48.9} & \scalebox{1}{53.3} & \scalebox{1}{46.6} & \scalebox{1}{52.8} & \scalebox{1}{52.2} & \scalebox{1}{55.6} & \scalebox{1}{53.9} & \scalebox{1}{56.7} &\scalebox{1}{53.3} &\scalebox{1}{53.3} &\scalebox{1}{50.6} &\scalebox{1}{57.2} &\scalebox{1}{54.4} & \scalebox{1}{55.0} &  \scalebox{1}{56.1} & \scalebox{1}{50.5} &\scalebox{1}{51.1} & \scalebox{1}{57.2} & \scalebox{1}{61.7}\\
    \scalebox{1}{SA} & \scalebox{1}{99.0} & \scalebox{1}{96.3} & \scalebox{1}{69.6} & \scalebox{1}{71.2} & \scalebox{1}{31.9} & \scalebox{1}{100.0} & \scalebox{1}{100.0} & \scalebox{1}{95.6} & \scalebox{1}{98.4} & \scalebox{1}{97.0} &\scalebox{1}{100.0} &\scalebox{1}{99.6} &\scalebox{1}{100.0} &\scalebox{1}{100.0} &\scalebox{1}{100.0} & \scalebox{1}{100.0} &  \scalebox{1}{98.8} & \scalebox{1}{81.4} &\scalebox{1}{100.0} & \scalebox{1}{99.0} & \scalebox{1}{99.9}\\
    \scalebox{1}{UG} &\scalebox{1}{93.8}  & \scalebox{1}{90.3} & \scalebox{1}{75.9} & \scalebox{1}{94.4} & \scalebox{1}{41.2} & \scalebox{1}{87.8} & \scalebox{1}{85.9} & \scalebox{1}{88.4} & \scalebox{1}{85.6} & \scalebox{1}{85.6} &\scalebox{1}{85.6} &\scalebox{1}{83.4} &\scalebox{1}{85.9} &\scalebox{1}{87.5} &\scalebox{1}{85.3} & \scalebox{1}{85.0} &  \scalebox{1}{86.6} & \scalebox{1}{82.1} &\scalebox{1}{80.3} & \scalebox{1}{85.3} &  \scalebox{1}{91.3}\\
    \midrule
    \scalebox{1}{Avg.}  & \bf80.05 & \scalebox{1}{67.0} & \scalebox{1}{66.0} & \scalebox{1}{72.5} & \scalebox{1}{48.6} & \scalebox{1}{71.8} & \scalebox{1}{70.9} & \scalebox{1}{70.3} & \scalebox{1}{71.9} & \scalebox{1}{71.5} &\scalebox{1}{72.1} &\scalebox{1}{70.8} &\scalebox{1}{71.1} &\scalebox{1}{72.7} &\scalebox{1}{70.7} & \scalebox{1}{71.0} &  \scalebox{1}{73.0}& \scalebox{1}{67.5} &\scalebox{1}{70.4} & \scalebox{1}{73.6} & \underline{\scalebox{1}{76.04}}\\
    \midrule
    \scalebox{1}{Rank}  & \bf4.35 & \scalebox{1}{15.20} & \scalebox{1}{15.55} & \scalebox{1}{10.25} & \scalebox{1}{19.55} & \scalebox{1}{10.40} & \scalebox{1}{11.70} & \scalebox{1}{12.40} & \scalebox{1}{9.40} & \scalebox{1}{9.95} &\scalebox{1}{8.90} &\scalebox{1}{12.80} &\scalebox{1}{11.50} &\scalebox{1}{7.30} &\scalebox{1}{12.40} & \scalebox{1}{12.85} &  \scalebox{1}{6.45}& \scalebox{1}{14.60} &\scalebox{1}{12.65} & \underline{\scalebox{1}{6.40}} & \underline{\scalebox{1}{6.40}}\\
	\bottomrule
  \end{tabular}
  }
    \end{small}
  \end{threeparttable}
\end{table}

In this section, we present the results of our experiments on benchmark datasets for time series classification tasks. We evaluate the performance of our proposed method, TSCMamba, and compare it with SOTA baseline models. Our method demonstrates superior performance across datasets of varying complexities. For example, TSCMamba outperforms many baseline methods on datasets with irregular patterns (e.g., EC and HW) in a large margin, indicating its ability to model 
 complex patterns effectively. Furthermore, our approach maintains strong performance even on datasets where traditional models, such as DTW and XGBoost, have historically performed well, demonstrating its versatility. Through comprehensive comparisons with SOTA baselines, we validate the effectiveness of TSCMamba in handling diverse time series classification challenges.
\subsection{Datasets}
We evaluated the performance of our proposed method, TSCMamba, on 10 benchmark datasets for time series classification tasks following TimesNet~\citep{wu2022timesnet} and an additional 20 datasets following TSLANet~\citep{tslanet} and UEA archive~\cite{bagnall2018uea}. These datasets are commonly used in the literature and are representative of various domains, including image, audio, and sensor data. We present dataset statistics in S-Table~\ref{tab:datasets} with more details in Appendix Subsection~\ref{sec:sup_dataset}. By evaluating our method on all 30 UEA datasets, we ensure a comprehensive assessment across diverse real-world time series classification challenges. This expansion strengthens the generalizability of our findings and aligns with prior SOTA benchmarks. These datasets are sourced from a diverse set of domains and contain a diverse range of classes, channels, and time-sequences leading to a robust benchmark for evaluating classification tasks. Moreover, some datasets contain more data in the Test set than the Train set (EC, HW, HB, JV, SCP1, UG) making the time-series classification a harder task. More domain-related information can be found in~\cite{bagnall2018uea}.

\subsection{Experimental Environment}
All experiments were conducted using the PyTorch framework~\citep{Paszke2019PyTorchAI} with NVIDIA 4$\times$ V100 GPUs (32GB each). The model was optimized using the ADAM algorithm~\citep{kingma2014adam} with Cross-Entropy loss following TimesNet~\citep{wu2022timesnet}. Moreover, the baseline results are utilized from TimesNet~\citep{wu2022timesnet} paper for a fair comparison (Same train-test set across the methods). The batch size, epochs, and initial learning rate varied on the datasets for optimal performance. The hyperparameters were selected based on the Train and Test set provided by the dataset archive following TimesNet~\cite{wu2022timesnet}. Moreover, the optimization was performed utilizing a cosine-annealing learning rate scheduler. We measure the prediction performance of our method using accuracy metric where larger values indicate better prediction accuracy.\\
\paragraph{Baseline Models}
In this study, we evaluate the performance of our proposed method, TSCMamba, against 20 state-of-the-art baselines in Table~\ref{tab:full_classification_results}, encompassing Transformer-based~\citep{tslanet,wu2022timesnet,NIPS2017_3f5ee243,kitaev2020reformer,haoyietal-informer-2021,liu2021pyraformer,wu2021autoformer,Liu2022NonstationaryTR,zhou2022fedformer,woo2022etsformer,wu2022flowformer}, CNN-based~\citep{Franceschi2019UnsupervisedSR}, RNN-based~\citep{Hochreiter1997LongSM,2018Modeling,gu2022efficiently}, MLP-based~\citep{Zeng2022AreTE,Zhang2022LessIM}, and classical machine learning-based methods~\citep{Berndt1994UsingDT,Chen2016XGBoostAS,Dempster2020ROCKETEF}. Therefore the comparison among these methods following TimesNet~\citep{wu2022timesnet} provides strong recent baselines from various aspects of machine learning.
\subsection{Predictive Performance Comparison}
The comprehensive results are presented in Table~\ref{tab:full_classification_results}. Notably, our approach achieves a substantial improvement of 4.01\% over the existing best baseline, TSLANet~\citep{tslanet}. Additionally, it improves upon the existing 
second-best baseline, TimesNet~\citep{wu2022timesnet}, by 6.45\%, which is a large margin compared to TimesNet's improvement of 0.6\% over the previous best baseline, Flowformer~\citep{wu2022flowformer}. This notable performance gain establishes TSCMamba as a strong contender for the TSC task. Unlike prior incremental improvements in TSC models, the large margin of improvement highlights the effectiveness of our multi-view fusion and tango scanning strategies, reinforcing their impact on long-range dependency modeling. For more insights regarding our scanning mechanism refer to~\ref{sub:explain_tango}. Our implementation is available at this \href{https://github.com/Atik-Ahamed/TSCMamba}{link}. 
 
\begin{figure}[H]
    \centering
    \includegraphics[width=\linewidth]{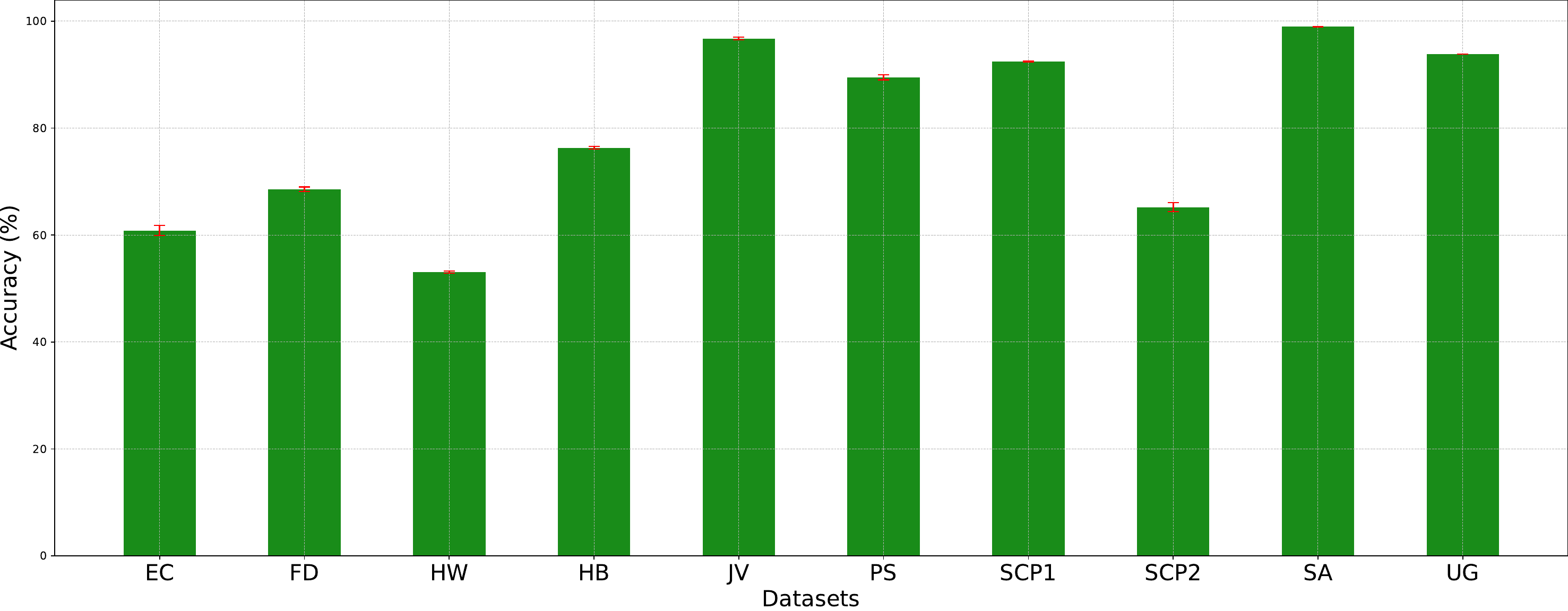}
    \caption{Performance of TSCMamba over 5 random runs. The mean performance is shown as green bars, with the standard deviation represented by red error bars that are very small. (Best viewed when zoomed in.)}
    \label{fig:error_bar}
\end{figure}

 While Table~\ref{tab:full_classification_results} presents our best-achieved results, we also demonstrate TSCMamba's reproducibility and stability across 5 runs with mean and error bars (standard deviation) in S-Figure~\ref{fig:error_bar}. In addition to the main baselines, we also compare TSCMamba against regular Mamba (Figure~\ref{fig:direct_mamba}). TSCMamba consistently outperforms regular Mamba, showcasing the effectiveness of our modifications. Specifically, the integration of tango scanning enhances inter-token dependencies, while multi-view fusion allows the model to adaptively leverage both spectral and temporal features. Beyond outperforming regular Mamba, TSCMamba also demonstrates superior performance compared to CNN-based methods such as TCN, which struggle to maintain sequence-level contextual information. This is evident in datasets where temporal dependencies play a crucial role, further validating the necessity of long-sequence modeling capabilities in TSC tasks.
 From Table~\ref{tab:full_classification_results}, it is evident that some methods may perform well on certain datasets (e.g., TimesNet~\citep{wu2022timesnet} on JV, SA), but may lack performance by a large margin on others (e.g., TimesNet~\citep{wu2022timesnet} on EC, HW). In contrast, our method maintains a balance across the datasets while showing a significant improvement in average performance.

In Table~\ref{tab:full_classification_results}, we compared our approach with various baselines including traditional methods, where TSLANet~\cite{tslanet} achieved the second-best performance. To further validate our model and findings, we evaluated an additional 20 datasets in Table~\ref{tab:additional_results} using the recent baselines from TSLANet.  Following TimesNet~\citep{wu2022timesnet} and other recent baselines like TSLANet~\citep{tslanet}, we present the results as average performance across datasets. However, recognizing that averaging over various datasets might not be the optimal way to assess different models, we also present ranks of different models as suggested by~\citet{JMLR:v7:demsar06a}. These ranks are calculated following the methodology of~\citet{IsmailFawaz2018deep}. This ranking method ensures a more robust comparison by normalizing performance across datasets of varying difficulty levels, providing a clearer picture of relative model performance than averaged accuracy values alone. For both sets of benchmark datasets, Table~\ref{tab:full_classification_results} and Table~\ref{tab:additional_results} demonstrate that TSCMamba achieves the best performance in terms of averaged accuracy and rank. Additionally, we provide Critical Difference (CD) diagrams in Appendix Subsection~\ref{sub:cd_diagrams} to visually compare model rankings across datasets. The diagrams illustrate the relative positioning of TSCMamba alongside other baselines, reinforcing its strong ranking across both benchmark and additional datasets. These visualizations complement our tabular results, providing an alternative perspective on model performance.

\begin{table}[H]
    \centering
    \caption{Additional classification results on the UEA datasets in terms of accuracy (as \%). The ranks are computed by averaging dataset-wise rankings of classifiers (lower is better).}
    \label{tab:additional_results}
    \resizebox{\linewidth}{!}{
    \begin{tabular}{lcccccccccc}
    \toprule
        Dataset & TSCMamba & TSLANet & GPT4TS & TimesNet & ROCKET & CrossF. & PatchTST & MLP & TS-TCC & TS2VEC\\ 
        &(Ours)&\citeyearpar{tslanet}&\citeyearpar{gpt4ts}&\citeyearpar{wu2022timesnet}&\citeyearpar{Dempster2020ROCKETEF}&\citeyearpar{zhang2023crossformer}&\citeyearpar{Yuqietal-2023-PatchTST}&\citeyearpar{Zeng2022AreTE}&\citeyearpar{tstcc}&\citeyearpar{ts2vec}\\
        \midrule

AtrialFibrillation  &  67.00 & 40.00  &  33.33  &  33.33  &  20.00  &  46.66  & 53.33 &  46.66  &  33.33  & 53.33 \\
BasicMotions  & 100.00 &100.00&  92.50  & 100.00 &  100.00  &  90.00  &  92.50  &  85.00  &  100.00  &  92.50  \\
Cricket & 98.61 & 98.61 &  8.33  &  87.50  & 98.61 &  84.72  &  84.72  &  91.67  &  93.06  &  65.28  \\
FingerMovements  & 69.00 &61.00  &  57.00  &  59.38  &  61.00  & 64.00 &  62.00  & 64.00 &  44.00  &  51.00  \\
HandMovementDirection  & 71.62 &52.70  &  18.92  &  50.00  &  50.00  & 58.11 &  58.11  &  58.11  & 64.86 &  32.43  \\

MotorImagery  & 62.00& 62.00 &  50.00  &  51.04  &  53.00  &61.00 &  61.00  &  61.00  &  47.00  &  47.00  \\

PenDigits  & 98.54 & 98.94 &  97.74  &  98.19  &  97.34  &  93.65  & 99.23 &  92.94  &  98.51  &  97.40  \\
PhonemeSpectra  & 24.66 & 17.75  &  3.01  & 18.24 &  17.60  &  7.55  &  11.69  &  7.10  & 25.92 &  8.23  \\
RacketSports  & 91.45& 90.79 &  76.97  &  82.64  & 86.18 &  81.58  &  84.21  &  78.95  &  84.87  &  74.34  \\
StandWalkJump  &  73.33 & 46.67  &  33.33  &  53.33  &  46.67  &  53.33  & 60.00 & 60.00 &  40.00  &  46.67  \\
InsectWingbeat&14.30& 10.00 & 10.00 & 10.00 & 10.00 & 10.00 & 10.00 & 10.00 & 10.00 & 10.00\\
DuckDuckGeese &62.00&24.00&50.00&56.00&52.00&44.00&24.00&62.00&38.00&68.00\\

NATOPS&94.44&95.56&91.67& 81.82&83.33&88.33&96.67&93.89&96.11& 82.22\\
Libras& 90.00&  92.78 & 79.44 & 77.84 & 83.89 & 76.11 & 81.11 & 73.33 & 86.67 & 85.56\\
ArticularyWordRecognition&97.00& 99.00& 93.33& 96.18& 99.33& 98.00& 97.67& 97.33 &98.00& 87.33\\
Epilepsy &  97.10 & 98.55 & 85.51 & 78.13 & 98.55 & 73.19 & 65.94 & 60.14 & 97.10 & 62.32\\
LSST& 60.30 &  66.34 & 46.39 & 59.21 & 54.10 & 42.82 & 67.80 & 35.77 & 49.23 & 39.01\\
ERing&91.50&95.56&95.93&94.07&94.07&94.07&92.59&91.48&90.40&87.40\\
CharacterTrajectories&99.00&98.75&98.26&97.98&98.75&98.33&97.35&97.35&98.50&99.50\\
EigenWorms& 87.00&41.22&48.09&58.78&78.63&54.96&54.20&41.22&77.90&84.70\\
\midrule
Average &\bf77.44& \underline{69.51}&58.49&67.18& 69.15&66.02& 67.71&65.40&68.67& 63.71 \\
\midrule
Rank &\bf 2.48&\underline{3.95}&7.40&6.00&4.95&6.17&5.12&6.80&5.22&6.90\\ 
\bottomrule
    \end{tabular}
    }
\end{table}

\subsection{Computational Complexity}
In this study, we compared the floating-point operations (FLOPs) (in Table~\ref{tab:flops}) of the top-performing methods (see Table~\ref{tab:full_classification_results}). To calculate FLOPs, we set a batch size of 16 across all baselines. For our method, we employed the best-performing hyperparameters, whereas for other baselines, we utilized the recommended parameters specified in the official TimesNet code~\citep{wu2022timesnet} and Flowformer~\citep{wu2022flowformer}. We leveraged the source code from~\cite{calflops} to calculate FLOPs. The overall FLOPs, including both forward and backward passes, are presented in Table~\ref{tab:flops}. Notably, our method achieves substantial improvements in terms of FLOPs across all datasets, with the exception of PEMS-SF (PS). This anomaly can be attributed to the projected space ($X$) used to achieve the best result for this dataset, which was set to 1024, thereby impacting the total FLOPs for this dataset only.
\begin{table}[t]
\caption{FLOPs comparison among the top performing methods. The values are represented in GigaFLOPS (G) or TeraFlops (T), where 1TFLOPs=1000GFLOPs and a lower value indicates better computational efficiency.}
\vskip 0.05in
    \centering
      \resizebox{\linewidth}{!}{
    \begin{tabular}{lcccccccccc}
    \toprule
        Methods&EC&FD&HW&HB&JV&PS&SCP1&SCP2&SA&UG\\
        \midrule
        Flow.~\cite{wu2022flowformer}&1.06T&37.97G&92.21G&246.37G&15.76G& 94.02G& 542.64G&697.74G&50.33G&190.82G\\
        TimesNet.~\cite{wu2022timesnet}&1.11T&161.93G&115.88G&182.69G&48.15G&\bf74.18G&503.62G&2.33T& 26.00G& 247.73G\\
        TSCMamba (Ours)&\bf1.69G&\bf11.53G&\bf27.24G&\bf8.39G&\bf12.33G&2.84T&\bf3.42G&{\bf{11.11}}G&\bf0.78G&\bf13.86G\\

        \bottomrule
    \end{tabular}
    }
    \label{tab:flops}
\end{table}

\section{Ablation}
\subsection{Component-wise Ablation}

We conducted an ablation study to investigate the contribution of individual components in our proposed method. The results are presented in Table~\ref{tab:components}. Each component is systematically removed or replaced to analyze its effect on classification accuracy.

Performance decreases significantly across all datasets when Mamba modules are removed, suggesting that these modules play a crucial role in our approach's effectiveness. Specifically, Mamba's state-space formulation allows it to capture long-range dependencies efficiently, and its removal leads to a notable drop in performance highlighting its importance. When Mamba is not employed (2nd row), the intermediate values are bypassed by scanning operations and directly fed into the DP and MLPs for class logit prediction. This results in significantly reduced model capacity, particularly for datasets requiring long-range contextual understanding.

In the 3rd row, we present depth-wise max-pooling instead of average-pooling, resulting in an input shape of B,3X. The decrease in accuracy across several datasets suggests that average pooling provides better feature aggregation by reducing variance and improving stability in feature representations.

Furthermore, when ROCKET-extracted features are not utilized (4th row), we resort to MLP-extracted features, where the former are non-learnable and the latter are learnable. The substantial accuracy drop indicates that ROCKET’s non-learnable convolutional features provide rich and robust representations that enhance classification. Additionally, in the 5th row, we explore the effect of replacing additive fusion (AF) with multiplicative fusion (MF) (of ROCKET features and spectral features), as detailed in Sec.\ref{sec:fusion}. While multiplicative fusion offers interaction-based feature integration, the observed performance degradation suggests that additive fusion better preserves the scale and discriminative properties of individual features, leading to more effective multi-view representation learning.

Notably, while Table \ref{tab:components} largely mirrors the best performance reported in Table ~\ref{tab:full_classification_results} across most datasets, the SpokenArabicDigits (SA) dataset exhibits optimal performance when employing depth-wise max-pooling and MLP-based features. This suggests that for certain datasets, local feature representations may be more informative than global ones, reinforcing the need for dataset-dependent finetuning.

\begin{table}[H]
\caption{Ablation experiments on particular components in our method.}
\vskip 0.05in
    \centering
    \resizebox{\linewidth}{!}{
    \begin{tabular}{ccccccccccccccc}
    \toprule
        Mamba&Avg.Pool&ROCKET&AF&EC&FD&HW&HB&JV&PS&SCP1&SCP2&SA&UG& Avg.\\
        \midrule
        \cmark&\cmark&\cmark&\cmark&62.0&57.0&53.3&74.1&93.0&90.2&92.5&66.7&94.1&93.8&77.67\\
        \xmark&\cmark&\cmark&\cmark&33.1&63.2&34.1&73.2&85.4&81.5&86.7&57.2&74.0&89.1&67.75\\
        \cmark&\xmark&\cmark&\cmark&31.6&64.2&52.0&74.1&94.1&63.0&86.7&60.6&96.7&92.8&71.58\\
        \cmark&\cmark&\xmark&\cmark&31.6&69.4&24.8&76.6&97.0&87.3&91.8&58.3&97.6&86.2&72.06\\
        \cmark&\cmark&\cmark&\xmark&30.0&51.5&49.3&72.7&91.4&84.4&88.7&58.9&90.0&90.3&70.72\\
        \bottomrule
    \end{tabular}
    }
    \label{tab:components}
\end{table}
\subsection{Comparison with Mamba}
In addition to the robust baselines used in Tables~\ref{tab:full_classification_results} and~\ref{tab:additional_results}, we also evaluated the performance of directly using Mamba~\citep{gu2023mamba}. This experiment aims to assess whether our modifications—such as the fusion of spectral and temporal features and the introduction of tango scanning—provide tangible benefits over a standard Mamba implementation. 

For the 10 benchmark datasets in Table~\ref{tab:full_classification_results}, we implemented our approach that directly applied regular Mamba by processing the MTS data through Mamba modules with a standard scanning protocol. The processed features were then passed through two-stage MLPs for classification, following the architecture in Figure~\ref{fig:method}. This setup ensures a direct comparison between our TSCMamba approach and the original Mamba framework, isolating the impact of our enhancements. 

\begin{figure}[H]
    \centering
    \includegraphics[width=\linewidth]{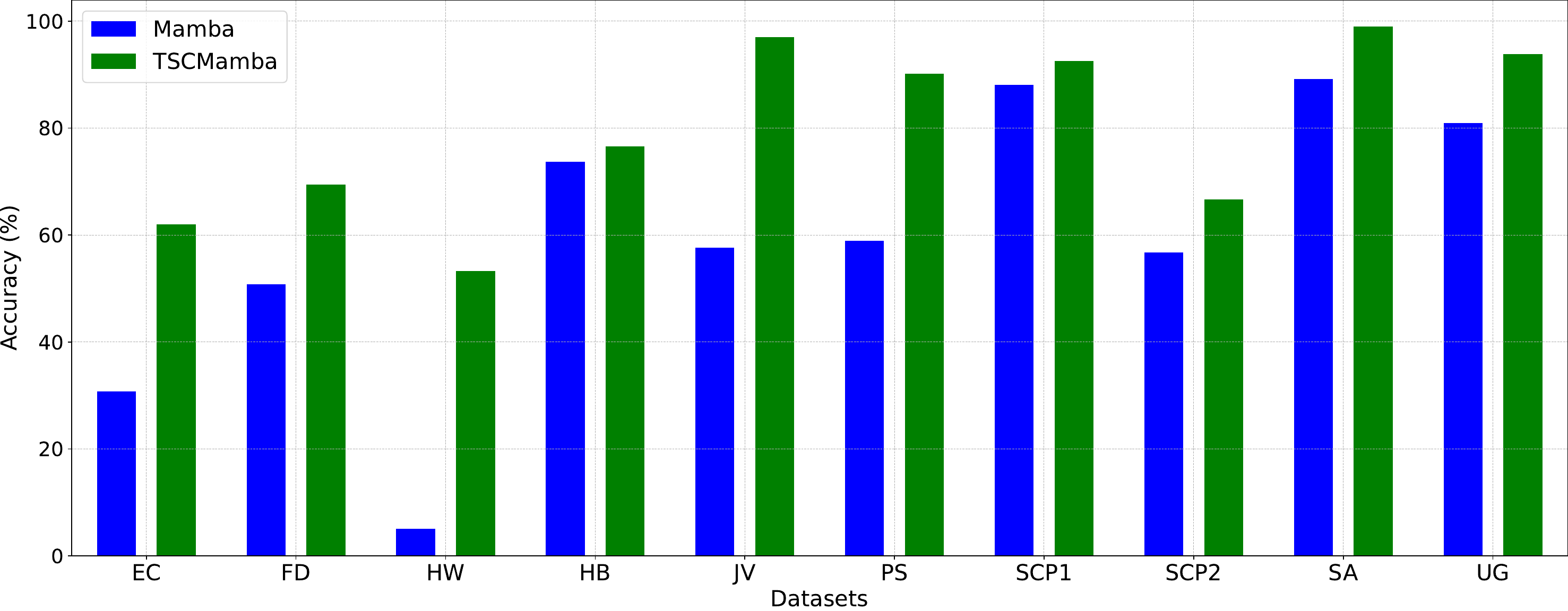}
    \caption{Comparison between TSCMamba and standard Mamba when applied to time series data.}
    \label{fig:direct_mamba}
\end{figure}

As shown in Figure~\ref{fig:direct_mamba}, TSCMamba achieves superior performance compared to all benchmarks across the 10 datasets, demonstrating that our modifications—particularly tango scanning and multi-view feature fusion—enhance the model’s ability to capture both local and global dependencies in MTS classification. All experiments maintain consistent hyperparameters with those reported in Table~\ref{tab:full_classification_results}, ensuring a fair evaluation where improvements can be attributed solely to the architectural enhancements rather than differences in hyperparameter tuning. 

\subsection{Comparison with BiMamba}
In this section, we compare TSCMamba's channel and token (or time) tango scanning (Eqs. \ref{eq_tango_channel} and \ref{eq_tango_time}) against BiMamba~\citep{schiff2024caduceus}. To ensure a fair comparison, we maintained the same hyperparameters used to achieve the results in Table~\ref{tab:full_classification_results}. Both models were trained under identical settings, including learning rates, batch sizes, and sequence lengths, ensuring that any performance differences arise from the architectural modifications rather than hyperparameter tuning. For BiMamba, we used the official code provided by~\citet{schiff2024caduceus}, with tied weights for the in and out projections. 

\begin{figure}[H]
    \centering
    \includegraphics[width=\linewidth]{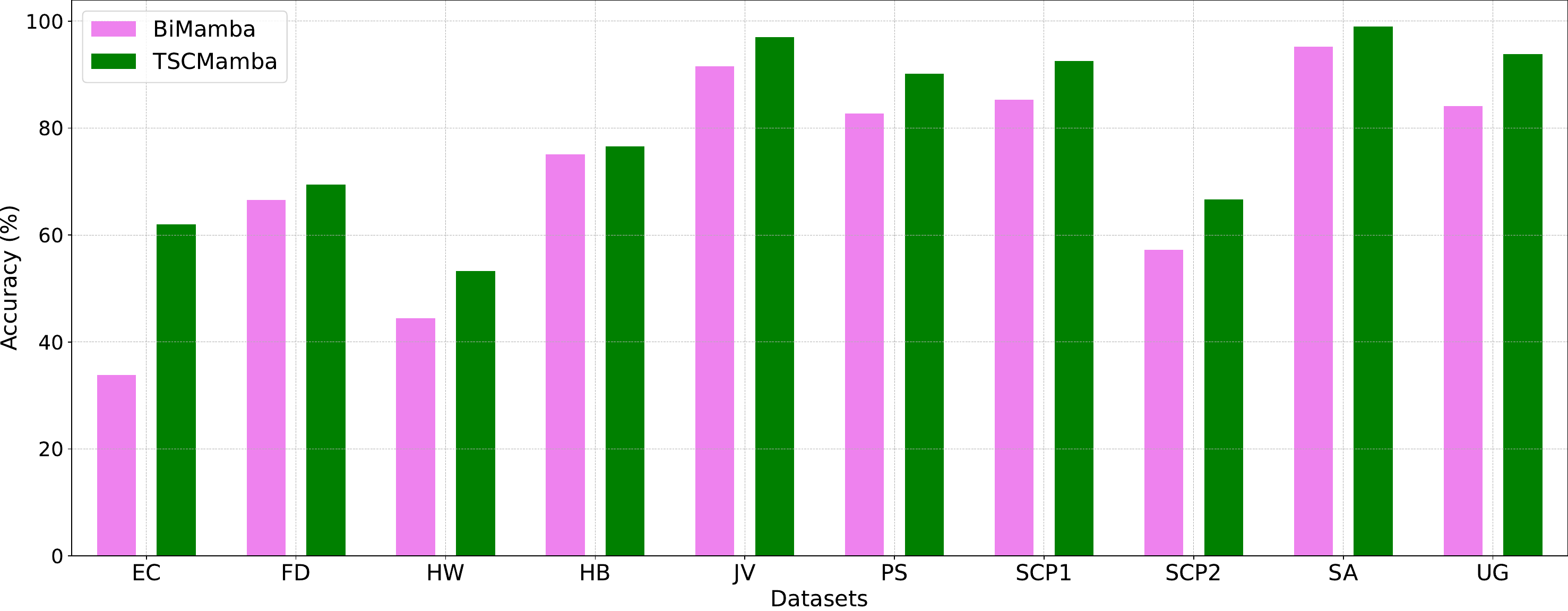}
    \caption{Performance comparison in accuracy of TSCMamba with tango scanning versus BiMamba~\citep{schiff2024caduceus} across the benchmark datasets. }
    \label{fig:bimamba}
\end{figure}

As illustrated in Figure~\ref{fig:bimamba}, TSCMamba consistently outperforms BiMamba across all benchmark datasets, demonstrating the effectiveness of tango scanning. The observed improvements indicate that tango scanning more effectively models inter-token dependencies, maximizing pairwise token interactions (For details, please see Subsection~\ref{sub:explain_tango}) while maintaining computational efficiency.

\subsection{Effectiveness of Tango Scanning}
While our tango scanning approach may seem counterintuitive compared to purely forward-based scanning or bidirectional scanning with reverse flips~\citep{schiff2024caduceus}, it achieves substantial improvements in accuracy. Unlike standard bidirectional approaches that treat both directions symmetrically, tango scanning strategically employs inversion invariance via flipping the sequence only once, enhancing inter-token interactions while maintaining computational efficiency. This strategy provides more complementary information for time series classification task. We provide more conceptual explanation in Subsection~\ref{sub:explain_tango}.

\begin{figure}[H]
    \centering
    \includegraphics[width=0.7\linewidth]{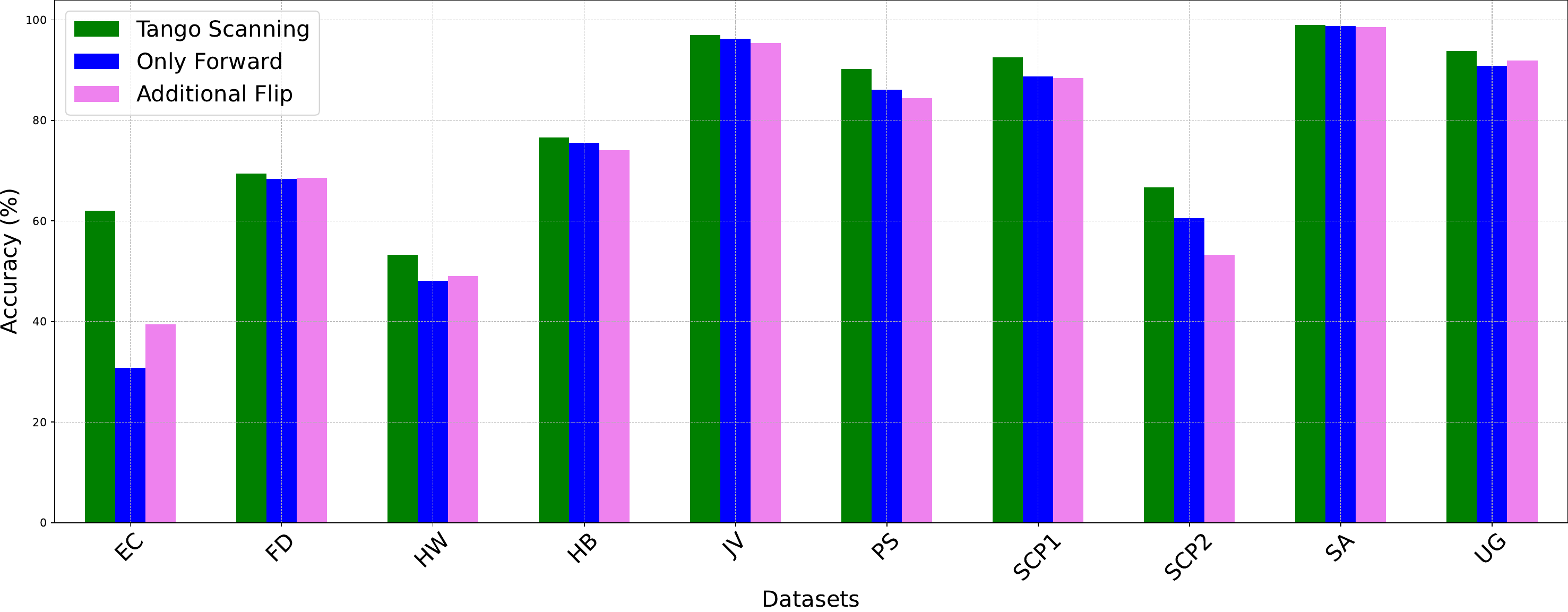}
    \caption{Effectiveness of our tango scanning compared against only forward-based scanning protocol and additional flip-based scanning protocol in reverse scanning.}
    \label{fig:tango_scanning}
\end{figure}

As shown in Figure~\ref{fig:tango_scanning}, this approach outperforms both traditional forward scanning and reverse-flip-based scanning methods on TSC tasks, particularly in complex scenarios. This improvement is especially notable in datasets where long-range dependencies are critical, as tango scanning enables more effective information flow across distant tokens.

We provide a theoretical explanation of tango scanning in Section~\ref{sec:tango_scaning}. These results highlight the advantages of our approach in effectively modeling sequence-level relationships while avoiding the redundancy and computational overhead typically associated with full bidirectional processing. 

\section{Hyperparameter Sensitivity}
\begin{figure}[H]
    \centering
    {{\includegraphics[width=0.24\linewidth]{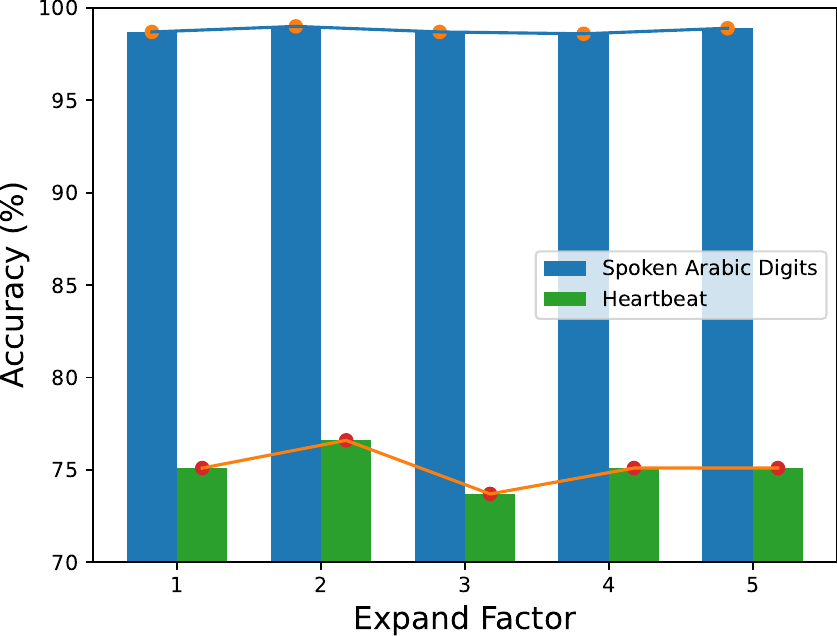} }}%
    {{\includegraphics[width=0.24\linewidth]{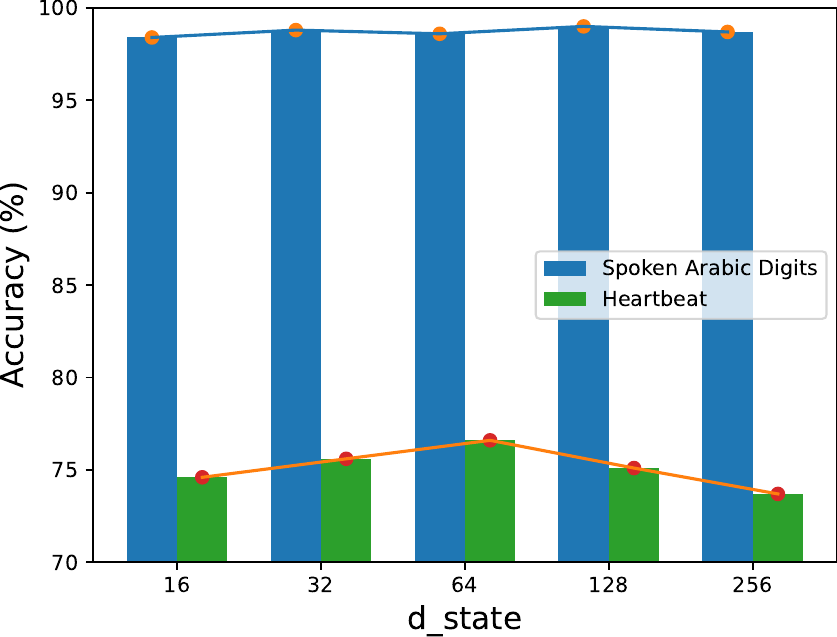} }}%
     {{\includegraphics[width=0.24\linewidth]{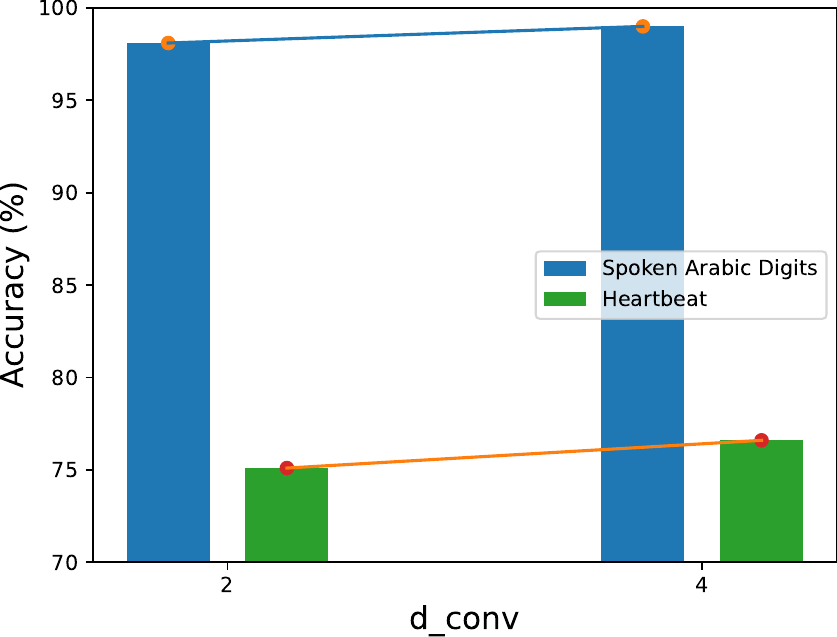} }}%
     {{\includegraphics[width=0.24\linewidth]{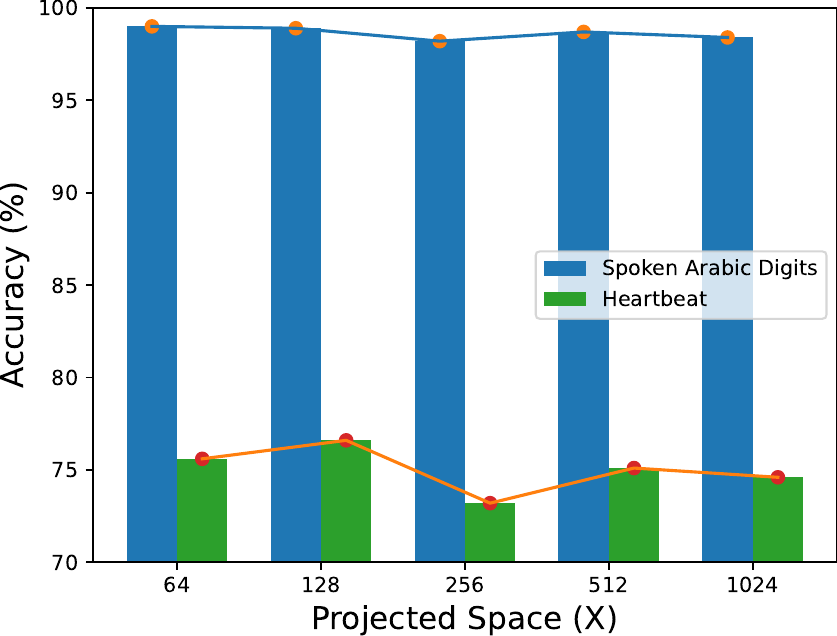} }}%
     
    \caption{Sensitivity analysis of TSCMamba's hyper-parameters on Time Series Classification (TSC) performance. The plot shows the impact of varying (from left to right, top to bottom) block expansion factor, SSM state expansion factor, local convolutional width, and dimension of the projected space (X) on model performance, highlighting the relative importance of each component in achieving optimal TSC results.}%
    \label{fig:mamba_ablation}%
\end{figure}

In this section, we discuss the key settings for the Mamba model~\citep{gu2023mamba}. The model operates with four main settings: model dimension (d\_model), SSM state expansion factor (d\_state), local convolution width (d\_conv), and block expansion factor (expand). While we derived the model dimension automatically from the input data dimensions, we tuned the remaining three parameters. Figure~\ref{fig:mamba_ablation} shows how these parameters affect model performance. In addition to Mamba's hyperparameters, we also tuned the dimension of the projected space ($X$) mentioned in~\ref{fig:method}.

\section{Conclusion and Future Work}
We present TSCMamba, an innovative approach for Time Series Classification (TSC) that enhances performance while reducing Floating Point Operations (FLOPs). By leveraging multi-view learning, TSCMamba analyzes time-series data through both local and global features extracted from time and frequency domains, effectively and efficiently capturing long-range dependent discriminative patterns in real-world data. Our novel tango scanning mechanism, validated through extensive experiments, demonstrates superior performance compared to conventional scanning methods.

Comprehensive evaluations show that TSCMamba consistently outperforms current state-of-the-art methods in both accuracy and computational efficiency across diverse TSC applications. However, our approach has some limitations. The fusion stage produces a tensor of $BD3X$ dimensions, where the $3X$ factor may increase computational requirements. Additionally, while we currently use CWT for spectral domain features, other transformations might provide complementary information that could further improve performance.

Future work will explore three key directions: incorporating self-supervised learning techniques, extending the framework to multiple-task learning beyond classification, and evaluating TSCMamba's effectiveness on more complex time-series datasets from various domains.

\section{Acknowledgment}
This research is supported in part by the NSF under Grant IIS 2327113 and ITE 2433190 and the NIH under Grants R21AG070909 and P30AG072946. We would like to thank NSF for support for the AI research resource with NAIRR240219. We thank the University of Kentucky Center for Computational Sciences and Information Technology Services Research Computing for their support and use of the Lipscomb Compute Cluster and associated research computing resources. We also acknowledge and thank those who created, cleaned, and curated the datasets used in this study.
\bibliographystyle{elsarticle-num-names} 
 \bibliography{main}

\section{Appendix}
In the appendix section, we present additional supplementary materials including algorithms, dataset statistics, etc. To distinguish from original materials, we add the prefix \textbf{S-} to the supplementary materials.
\setcounter{table}{0}
\setcounter{algorithm}{0}
\captionsetup[table]{name={S-Table}}
\captionsetup[algorithm]{name={S-Algorithm}}
\captionsetup[figure]{name={S-Figure}}

\subsection{Algorithms}
In this section, we present the two schematic algorithms for conversion of raw signals to CWT and ROCKET feature extraction in S-Algorithm~\ref{alg:cwt_conversion} and S-Algorithm~\ref{alg:rocket} respectively.
\begin{algorithm}[H]
\caption{Convert Raw Signals to CWT representation}
\label{alg:cwt_conversion}
\begin{algorithmic}[1]
\Require Raw signals of shape $(N, D, L)$
\Ensure Tensor of shape $(N, D, L_1, L_1)$ \# We set $L_1 = 64$ in this paper. 
\For{each signal $i$ in $N$}
    \For{each dimension $d$ in $D$}
        \State $signal \gets \text{Raw}[i, d, :]$
        \State $coeff,freq \gets \text{CWT}(signal)$
        \State $cwt\_resized \gets \text{resize}(coeff, (L_1, L_1), \text{mode=``constant''})$
        \State $\text{Tensor}[i, d, :, :] \gets cwt\_resized$
    \EndFor
\EndFor
\end{algorithmic}
\end{algorithm}
\begin{algorithm}[H]
\caption{Feature Extraction with ROCKET for Random Convolutional Kernel Transform}
\label{alg:rocket}
\begin{algorithmic}[1]
\Require Time series data of length = $L$, number of kernels, $n=\frac{X}{2}$
\Ensure Feature vector of shape $(2\times n ) =X$
\State $kernels \gets $ list of $n$ random kernels of random length $l$, weight $w$, bias $b$, dilation $d$, padding $p$.
\State $feature\_maps \gets $ empty list
\For{each kernel $k$ in $kernels$}
\State $h_{ppv},h_{max} \gets \text{convolve}(k, x)$
\State $feature\_maps.\text{append}(h_{ppv},h_{max})$
\EndFor
\State \Return feature\_ maps
\end{algorithmic}
\end{algorithm}

\subsection{Ablation on Tango Scanning}
\begin{figure}[H]
    \centering
    \includegraphics[width=\linewidth]{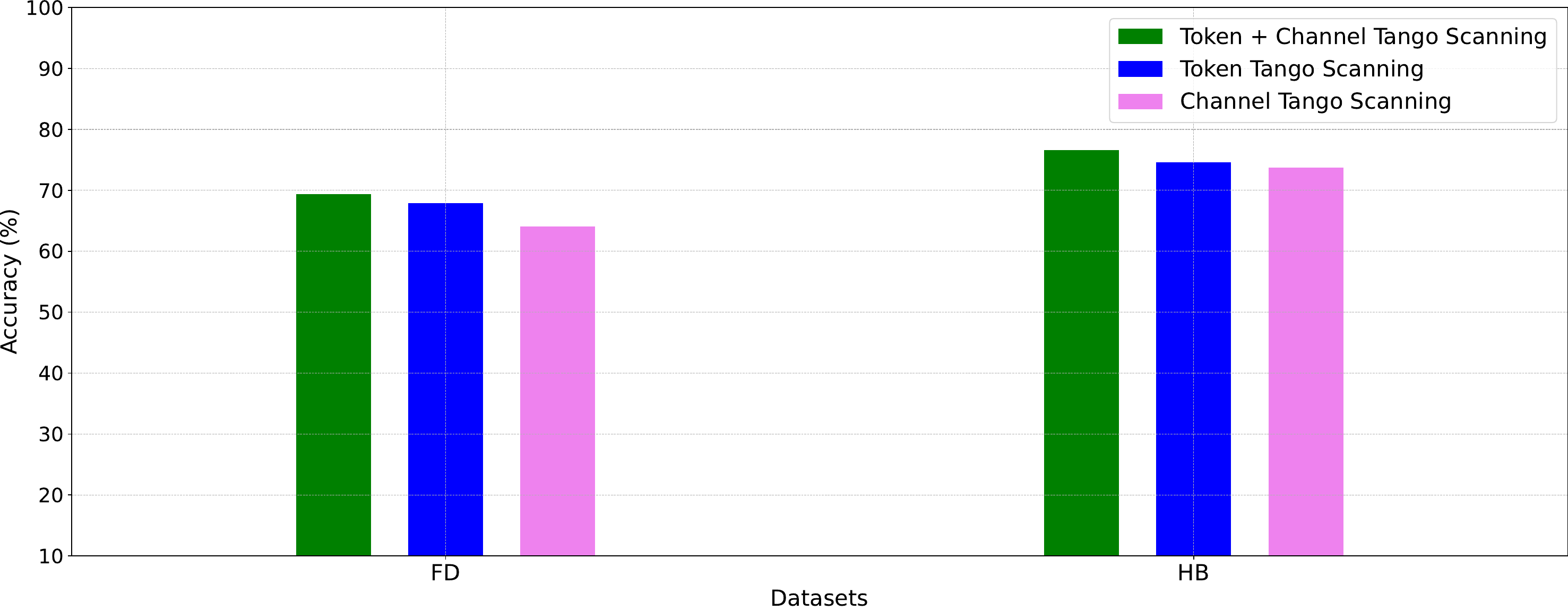}
    \caption{Ablation on tango scanning module compared against only token tango scanning and only channel tango scanning.}
    \label{fig:scanning_ablation}
\end{figure}

This section presents the results of our experiments with both channel and token tango scanning. We display the outcomes of token tango scanning with the channel tango scanning module turned off. Conversely, we also provide results for channel tango scanning with the token tango scanning module disabled. As demonstrated in S-Figure~\ref{fig:scanning_ablation}, both modules are crucial for achieving optimal accuracy, highlighting their individual importance in the overall performance of our model.
\subsection{Accuracy on Multiple Runs}

\subsection{Mamba’s Ability to Capture Relevant Information and Remote Dependencies}
\begin{figure}[H]
    \centering
    \includegraphics[width=\linewidth]{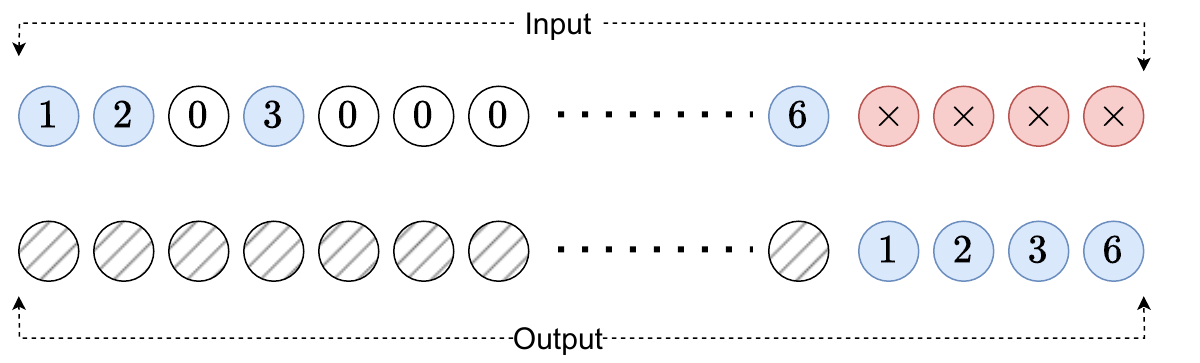}
    \caption{Conceptual diagram of selective copying experiment.}
    \label{fig:slcopy}
\end{figure}
Mamba’s ability to capture the most relevant information and model remote dependencies is enabled by its selective state-space mechanism, which dynamically parameterizes state transitions based on input content. This allows the model to filter out irrelevant inputs and selectively propagate key information, addressing a major limitation of prior models that rely on static time-invariant processing. To validate this, we conducted the selective copying Task, a synthetic benchmark designed to test content-aware recall by requiring the model to memorize randomly spaced relevant tokens while ignoring noise — a challenge for conventional models like CNNs or Linear Time-Invariant (LTI)-SSMs that lack content-awareness. An overview of the selective copying task is demonstrated in S-Figure~\ref{fig:slcopy}, which illustrates the design of the selective copying task, where the model must memorize specific tokens (light-blue) while ignoring irrelevant noise tokens (white). Unlike standard copying tasks with fixed spacing, this task introduces variable spacing, requiring content-aware selection for effective memorization. Following the original Mamba paper’s setup, we trained Mamba models on sequences of length 4096, using a 16-token vocabulary, and evaluated their ability to recall 16 relevant tokens while ignoring noise tokens. Our conducted results show that Mamba achieved \textbf{98.73\%} accuracy with 2 layers and \textbf{99.41\%} with 3 layers, demonstrating its ability to adaptively retain information across long sequences. These results confirm that Mamba effectively captures long-range dependencies by dynamically selecting relevant tokens, making it highly effective for tasks requiring precise memory retention and noise filtering, such as time series classification.

\subsection{Critical Difference Diagrams}
\label{sub:cd_diagrams}
\begin{figure}[H]
    \centering
    \includegraphics[width=\linewidth, trim={5cm 2cm 5cm 0cm}, clip]{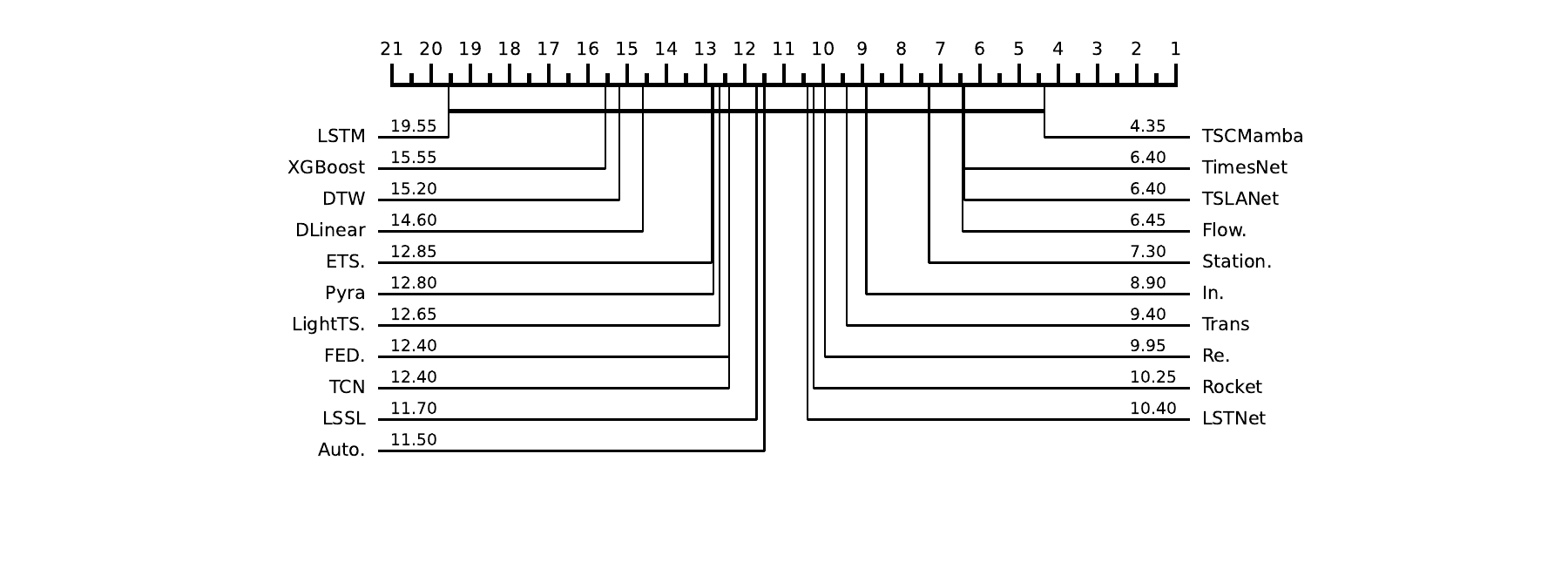}
    \caption{Critical difference diagram comparing performance differences between TSCMamba and baseline methods across 10 benchmark datasets. Lower mean ranks indicate better performance. 'Flow.' represents FlowFormer; other Transformer-based method names follow similar abbreviation patterns.}
    \label{fig:cd_benchmark}
\end{figure}

\begin{figure}[H]
    \centering
    \includegraphics[width=\linewidth, trim={5cm 0cm 5cm 0cm}, clip]{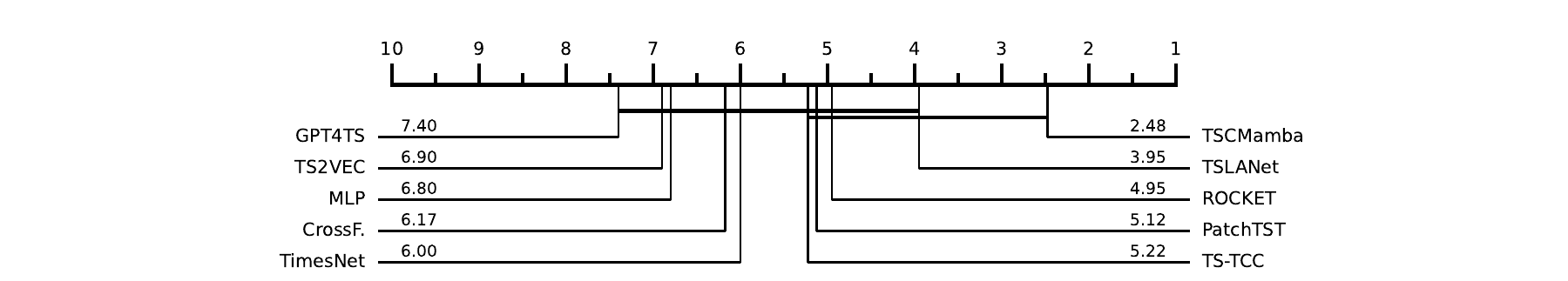}
    \caption{Critical difference diagram for the additional 20 datasets.}
    \label{fig:cd_additional}
\end{figure}

In this section, we present the critical difference diagrams in S-Figure~\ref{fig:cd_benchmark} and S-Figure~\ref{fig:cd_additional} following~\cite{Dempster2020ROCKETEF,shapelet}. We utilize the method from~\cite{IsmailFawaz2018deep} to generate the critical difference diagrams.
\subsection{Dataset Description}
\label{sec:sup_dataset}
We utilized all 30 datasets from the UEA classification archive~\cite{bagnall2018uea}. More specifically, the datasets' overview is mentioned below:\\
\textbf{EthanolConcentration (EC):} A spectral dataset of 44 whisky bottles with ethanol concentrations 35\%, 38\%, 40\%, and 45\%~\cite{large2018detecting}. Each sample has three repeat readings (226–1101.5nm) using a StellarNet BLACKComet-SR spectrometer. 28 standard bottles (clear, cylindrical) and 16 non-standard (irregular, colored) introduce noise. The leave-one-bottle-out evaluation ensures unseen test bottles. The dataset spans UV, VIS, and NIR ranges, with potential benefits from multivariate analysis.\\
\textbf{FaceDetection (FD):}Kaggle competition dataset~\cite{decoding-the-human-brain} with MEG recordings from 10 train subjects (subject 01-10) and 6 test subjects (subject 11-16). Each subject was recorded for 1.5 seconds (250Hz) with class labels Face (1) / Scramble (0). Data was high-pass filtered at 1Hz and preprocessed using mne-python.\\
\textbf{Handwriting (HW):} A smartwatch motion dataset reported in~\cite{shokoohi2017generalizing}, capturing accelerometer data while subjects write 26 letters. It includes 150 train cases and 850 test cases, with three dimensions representing accelerometer values. The data has been padded by the donors.\\
\textbf{Heartbeat (HB):} This dataset, from the PhysioNet/CinC Challenge 2016~\cite{physionet}, contains heart sound recordings from both healthy individuals and patients with cardiac conditions. Data was collected globally in clinical and non-clinical settings, from nine body locations, including the aortic, pulmonic, tricuspid, and mitral areas. Recordings are 5 seconds long and categorized as normal or abnormal, with abnormalities linked to heart valve defects and coronary artery disease. Spectrograms were generated using a 0.061s window with 70\% overlap, forming a multivariate dataset where each dimension represents a frequency band.\\
\textbf{JapaneseVowels (JV):} This UCI Archive dataset~\cite{kudo1999multidimensional} features recordings of nine Japanese male speakers pronouncing the vowels "a" and "e". A 12-degree linear prediction analysis transforms the raw audio into 12-dimensional time series of varying lengths (7 to 29), which were padded to length 29. The task is speaker classification (labels 1–9). The training set contains 30 utterances per speaker, while the test set varies between 24 and 88 utterances per speaker due to external factors.\\
\textbf{PEMS-SF (PS):} This dataset~\cite{cuturi2011fast} from the California Department of Transportation records 15 months of daily traffic occupancy rates (0 to 1) from San Francisco Bay Area freeways. Data spans Jan 1, 2008 – Mar 30, 2009, sampled every 10 minutes across 963 sensors. Each day forms a time series of length 144 ($6 \times 24$), with public holidays and anomalous days removed, resulting in 440 time series. The task is day-of-the-week classification (Monday–Sunday, labeled 1–7).\\
\textbf{SelfRegulationSCP1 (SCP1):} This BCI II dataset (Ia)~\cite{birbaumer1999spelling} from the University of Tuebingen records slow cortical potentials (SCPs) from a healthy subject controlling a cursor through self-regulation of brain activity. Cortical positivity moves the cursor down, while negativity moves it up. Each trial lasts 6s, with 3.5s (256Hz, 896 samples per channel) used for training/testing. The dataset contains 268 training trials (168 from day 1, 100 from day 2) and 293 test trials. Each instance has six EEG channels, and the classification task is to predict negativity or positivity.\\
\textbf{SelfRegulationSCP2 (SCP2):} This BCI II dataset (Ib)~\cite{birbaumer1999spelling} records slow cortical potentials (SCPs) from an artificially respirated ALS patient controlling a cursor through self-regulation of brain activity. Cortical positivity moves the cursor down, while negativity moves it up. Each trial lasts 8s, with 4.5s (256Hz, 1152 samples per channel) used for training/testing. The dataset includes 200 training trials (100 per class) and 180 test trials, all recorded on the same day. Data consists of 7 EEG channels, and the classification task is to predict positivity or negativity.\\
\textbf{SpokenArabicDigits (SA):} This UCI dataset contains 8,800 audio samples of spoken Arabic digits (0–9) from 88 native speakers (44 males, 44 females, aged 18–40). Each digit was recorded 10 times per speaker. The 13 Mel-Frequency Cepstral Coefficients (MFCCs) were extracted using a sampling rate of 11,025 Hz, 16-bit resolution, and Hamming window with pre-emphasis filtering.\\
\textbf{UWaveGestureLibrary (UG):} This dataset~\cite{liu2009uwave} contains eight simple gestures recorded using accelerometers. Each gesture is represented by x, y, and z coordinates, with a series length of 315.\\
\textbf{AtrialFibrillation (AF):} This two-channel ECG dataset~\cite{moody2004spontaneous} from the Computers in Cardiology Challenge 2004 is designed for predicting spontaneous termination of atrial fibrillation (AF). Each instance is a 5-second ECG segment with two signals sampled at 128 Hz. The dataset has three classes: n (AF that does not terminate for at least an hour), s (AF that self-terminates at least one minute after recording), and t (AF that terminates within one second of recording end). Each ECG channel is treated as a separate dimension in this multivariate dataset.\\
\textbf{BasicMotions (BM):} This dataset was collected in a 2016 student project, where four participants performed standing, walking, running, and playing badminton while wearing a smartwatch. The watch recorded 3D accelerometer and 3D gyroscope data at 10 Hz for 10 seconds per trial, with each participant repeating the activities five times.\\
\textbf{Cricket (CR):} This dataset~\cite{ko2005online,shokoohi2017generalizing} captures cricket umpire signals, where four umpires performed 12 distinct signals (e.g., No-Ball, TV Replay) 10 times each while wearing wrist-mounted accelerometers. Data was recorded at 184 Hz, with each accelerometer measuring x, y, and z axes, resulting in a six-dimensional time series from two sensors.\\
\textbf{FingerMovements (FM):} This EEG dataset~\cite{blankertz2001classifying} from Fraunhofer-FIRST, Intelligent Data Analysis Group (Klaus-Robert Mller) and Freie Universität Berlin records self-paced key typing from a subject typing $\approx~1$ key per second over three 6-minute sessions. Each trial captures 28 EEG channels for 500ms, ending 130ms before a key press, downsampled to 100 Hz (50 observations per channel). The dataset includes 316 training cases and 100 test cases, recorded using a NeuroScan amplifier and Ag/AgCl electrode cap following the 10/20 system.\\
\textbf{HandMovementDirection (HMD):} This Brain-Computer Interfaces (BCI) IV competition dataset records Magnetoencephalography (MEG) signals from two subjects moving a joystick (right, up, down, left) using only their hand and wrist after a prompt. The task is to classify the movement direction from 10 MEG channels over motor areas. Each instance captures data from 4s before to 6s after movement. \\
\textbf{MotorImagery (MI):} This dataset~\cite{lal2004methods} from the University of Tbingen records ECoG signals from a subject performing imagined movements of either the left small finger or the tongue. An 8×8 platinum electrode grid was placed on the right motor cortex, capturing 64-channel signals at 1000 Hz. Each 3-second trial starts 0.5s after a visual cue to avoid visual evoked potentials. The dataset includes 278 training cases and 100 test cases, with class labels finger or tongue movement.\\
\textbf{PenDigits (PD):} This is a handwritten digit classification task, where 44 writers drew digits 0–9 on a digital screen. Each instance consists of the x and y coordinates of the pen-tip movement, originally recorded at 500×500 resolution, then normalized to 100×100. The data~\cite{alimouglu2001combining} was spatially resampled to 8 points, resulting in a 2D time series of 8 points per instance, with a class label (0–9) indicating the drawn digit.\\
\textbf{PhonemeSpectra (PHS):} 
This multivariate dataset~\cite{hamooni2014dual} is derived from Google Translate audio recordings at 22,050 Hz, containing speech from male and female speakers. Waveforms were segmented into phonemes using the Forced Aligner tool from the Penn Phonetics Laboratory. Each instance is represented as a spectrogram with a 0.001s window and 90\% overlap, where each dimension corresponds to a frequency band. The dataset includes 39 classes.\\
\textbf{RacketSports (RS):} This dataset was collected from university students playing badminton or squash while wearing a Sony Smartwatch 3. The watch recorded x, y, z coordinates from both the gyroscope and accelerometer, transmitting data to a OnePlus 5 phone. The task is to classify the sport (badminton or squash) and stroke type (forehand/backhand for squash, clear/smash for badminton). Data was sampled at 10 Hz over 3 seconds as part of an undergraduate project (2017/18) by Phillip Perks.\\
\textbf{StandWalkJump (SWJ):} This PhysioNet dataset~\cite{behravan2015rate} records short-duration ECG signals from a 25-year-old healthy male performing standing, walking, and jumping to study motion artifacts. Data was sampled at 500 Hz (16-bit resolution) with an analog gain of 100. A spectrogram was generated using a 0.061s window with 70\% overlap, where each dimension represents a frequency band. The dataset has three classes (standing, walking, jumping), each with 9 instances.\\
\textbf{InsectWingbeat (IW):} This dataset~\cite{chen2014flying} reconstructs insect flight sounds captured by sensors. The dataset consists of power spectrum features derived from 1-second sound segments, with spectrograms generated using a 0.061s window and 70\% overlap. Each instance represents a frequency band from the spectrogram. The dataset has 10 classes, including male and female mosquitoes (Ae. aegypti, Cx. tarsalis, Cx. quinquefasciatus, Cx. stigmatosoma), two fly species (Musca domestica, Drosophila simulans), and other insects.\\
\textbf{DuckDuckGeese (DDG):} This dataset consists of bird recordings sourced from the Xeno Canto website, specifically from A and B quality categories. All recordings were downsampled to 44,100 Hz, truncated to 5 seconds, and converted into spectrograms using a 0.061s window with 70\% overlap. The dataset includes five bird species, each with 20 instances: Black-bellied Whistling Duck, Canadian Goose, Greylag Goose, Pink-footed Goose, and White-faced Whistling Duck.\\
\textbf{NATOPS (NP):} The NATOPS dataset~\cite{ghouaiel2017continuous}, from the AALTD 2016 competition, involves automatic detection of Naval Air Training and Operating Procedures Standardization (NATOPS) motions used to control aircraft. Data was recorded from sensors on the hands, elbows, wrists, and thumbs, capturing x, y, z coordinates at eight locations, resulting in 24 dimensions. The dataset includes six classes, representing different aircraft signaling motions: I have command, All clear, Not clear, Spread wings, Fold wings, and Lock wings.\\
\textbf{Libras (LB):} This represents Brazilian Sign Language (LIBRAS) hand movements as bi-dimensional curves~\cite{dias2016algoritmos}. It consists of 15 classes, each with 24 instances, recorded from four individuals over two sessions. Each 7-second video of a hand movement was processed to extract 45 frames, capturing the centroid of the hand in each frame. The data is normalized, and each movement is mapped to 90 features representing x and y coordinates.\\
\textbf{ArticularyWordRecognition (AWR):} The Electromagnetic Articulograph (EMA) dataset~\cite{shokoohi2017generalizing} captures tongue and lip movements during speech using an EMA AG500 system with 0.5 mm spatial accuracy. Data was collected from native English speakers pronouncing 25 words, using 12 sensors placed on the forehead, tongue (tip to back), lips, and jaw. Each sensor recorded x, y, and z positions at 200 Hz, resulting in 36 dimensions, though only 9 dimensions are included in this dataset. Head-mounted sensors were used to compute head-independent motion.\\
\textbf{Epilepsy (EP):} This dataset, presented in~\cite{villar2016generalized}, captures wrist-worn accelerometer data from six participants performing four activities: walking, running, sawing, and mimicked seizures. Each participant performed each activity at least 10 times, following a medical expert-defined protocol for seizure mimicking. Data was sampled at 16 Hz, with activities varying in length, but all were truncated to ~13 seconds for consistency. After data cleaning, 275 cases remained, with a train-test split of three participants each and IDs removed for consistency.\\
\textbf{LSST (LS):} The PLAsTiCC dataset, from a 2018 Kaggle competition~\cite{PLAsTiCC-2018}, is designed for classifying simulated astronomical time-series data in preparation for the Large Synoptic Survey Telescope (LSST) survey. It consists of light curves measuring an object's brightness over time across six passbands. The dataset includes multiple astronomical object classes, each driven by different physical processes. A 36-dimensional representation was chosen to minimize data truncation.\\
\textbf{ERing (ER):} This dataset was collected using a prototype finger ring with electric field sensing to detect hand and finger gestures~\cite{wilhelm2015ering}. It focuses on finger posture recognition with six classes involving the thumb, index, and middle fingers. Each instance is four-dimensional, with 65 observations per series, representing measurements from electrodes that vary based on hand proximity.\\
\textbf{CharacterTrajectories (CT):} Pen Tip Trajectory dataset, provided by Ben Williams (University of Edinburgh)~\cite{williams2007modelling}, contains 2,858 character samples recorded using a WACOM tablet. Each instance represents a 3D pen tip velocity trajectory (x, y, and force) captured at 200 Hz. Data was numerically differentiated, Gaussian smoothed ($\sigma=2$), normalized, and segmented based on pen tip force. Only single PEN-DOWN characters were included, with velocity profiles aligned to the dataset mean. The dataset consists of 20 character classes (a, b, c, ... z, excluding f, i, j, k, t, and x).\\
\textbf{EigenWorms (EW):} The C. elegans Movement dataset~\cite{yemini2013database} analyzes roundworm motion to study behavioral genetics. Worm movement is represented using six eigenworms, capturing each frame as a six-dimensional time series. The dataset classifies worms as wild-type (N2 strain) or four mutant types (goa-1, unc-1, unc-38, unc-63). Data was extracted from the C. elegans behavioral database.

Datasets with their corresponding number of channels ($D$), sequence length, train samples, test samples, number of classes, and domain information are presented in S-Table~\ref{tab:datasets}.
\begin{table}[H]
\caption{Summary statistics of publicly available datasets utilized in this paper.}
\label{tab:datasets}
\vskip 0.05in
    \centering
    \resizebox{\linewidth}{!}{
    \begin{tabular}{c|l|cccccc}
    \toprule
        &Datasets & Channels & Length & Train & Test & Classes & Domain\\
        \midrule
        \multirow{10}{*}{\rotatebox{90}{Benchmark datasets}}&
        EthanolConcentration (EC)& 3 & 1751& 261&263& 4& Alcohol Industry\\
        &FaceDetection (FD)& 144& 62&5890&3524&2&Face (250Hz)\\
        &Handwriting (HW)& 3 & 152& 150&850&26&Smart Watch\\
        &Heartbeat (HB)& 61& 405&204&205&2&Clinical\\
        &JapaneseVowels (JV)& 12& 29& 270 & 370&9& Audio\\
        &PEMS-SF (PS)& 963& 144& 267&173&7&Transportation\\
        &SelfRegulationSCP1 (SCP1)& 6& 896&268&293&2&Health (256Hz)\\
        &SelfRegulationSCP2 (SCP2)& 7& 1152&200&180&2&Health (256Hz)\\
        &SpokenArabicDigits (SA)&13&93&6599&2199&10&Voice (11025Hz)\\
        &UWaveGestureLibrary (UG)&3&315&120&320&8&Gesture\\
        \midrule
         \multirow{20}{*}{\rotatebox{90}{Additional datasets}}&
        AtrialFibrillation (AF)& 2 & 640& 15&15& 3& ECG\\
         &BasicMotions (BM)& 6& 100&40&40&4&Human Activity Recognition\\
        &Cricket (CR)& 6& 1197&108&72&12&Human Activity Recognition\\
        &FingerMovements (FM)&28&50&316&100&2&EEG\\
        &HandMovementDirection (HMD)&10&400&160&74&4&MEG\\
        &MotorImagery (MI)&64&3000&278&100&2&EEG\\
        &PenDigits (PD)&2&8&7494&3498&10&Motion\\
        &PhonemeSpectra (PHS)&11&217&3315&3353&39&Audio\\
        &RacketSports (RS)&6&30&151&152&4&Human Activity Recognition\\
        &StandWalkJump (SWJ)&4&2500&12&15&3&ECG\\
        &InsectWingbeat (IW)&200&variable&25000&25000&10&Bioacoustics\\
        &DuckDuckGeese (DDG)&1345&270&50&50&5&Bioacoustics\\
        &NATOPS (NP)&24&51&180&180&6&Human Activity Recognition\\
        &Libras (LB)& 2 &45&180&180&15&Sign Language Recognition\\
        &ArticularyWordRecognition (AWR)&9&144&275&300&25&Speech Articulation\\
        &Epilepsy (EP)&3&206&137&138&4&Human Activity Recognition\\
        &LSST (LS)&6&36&2459&2466&14&Astronomy\\
        &ERing (ER)&4&65&30&270&6&Human Activity Recognition\\
        &CharacterTrajectories (CT)&3&variable&1422&1436&20&Motion Analysis\\
        &EigenWorms (EW)&6&17984&128&131&5&Bioinformatics\\
        \bottomrule
    \end{tabular}
    }
 \end{table}   

\end{document}